%% file: main.tex
\documentclass[10pt,twocolumn,letterpaper]{article}

 \usepackage[pagenumbers]{cvpr} % To force page numbers, e.g. for an arXiv version

\usepackage{blindtext}
\usepackage{graphicx}
\usepackage[accsupp]{axessibility}

\input{preamble}

\definecolor{cvprblue}{rgb}{0.21,0.49,0.74}
\usepackage[pagebackref,breaklinks,colorlinks,citecolor=cvprblue]{hyperref}

\def\paperID{15182} % *** Enter the Paper ID here
\def\confName{CVPR}
\def\confYear{2024}

\title{Your Student is Better Than Expected: \\Adaptive Teacher-Student Collaboration for Text-Conditional Diffusion Models}

\author{Nikita Starodubcev \\
\and
Artem Fedorov
\and
Artem Babenko
\and
Dmitry Baranchuk
\\ 
{\hspace{-120mm}Yandex Research} \\ 
{\hspace{-120mm}\small{\url{https://github.com/yandex-research/adaptive-diffusion}}}
}

\newcommand{\dmitry}[1]{{\color{red}[\textbf{Dmitry}:#1]}}

\begin{document}
\twocolumn[{%
\renewcommand\twocolumn[1][]{#1}%
\maketitle
\begin{center}
    \centering
    \captionsetup{type=figure}
    \vspace{-1mm}
    \includegraphics[width=0.97\linewidth]{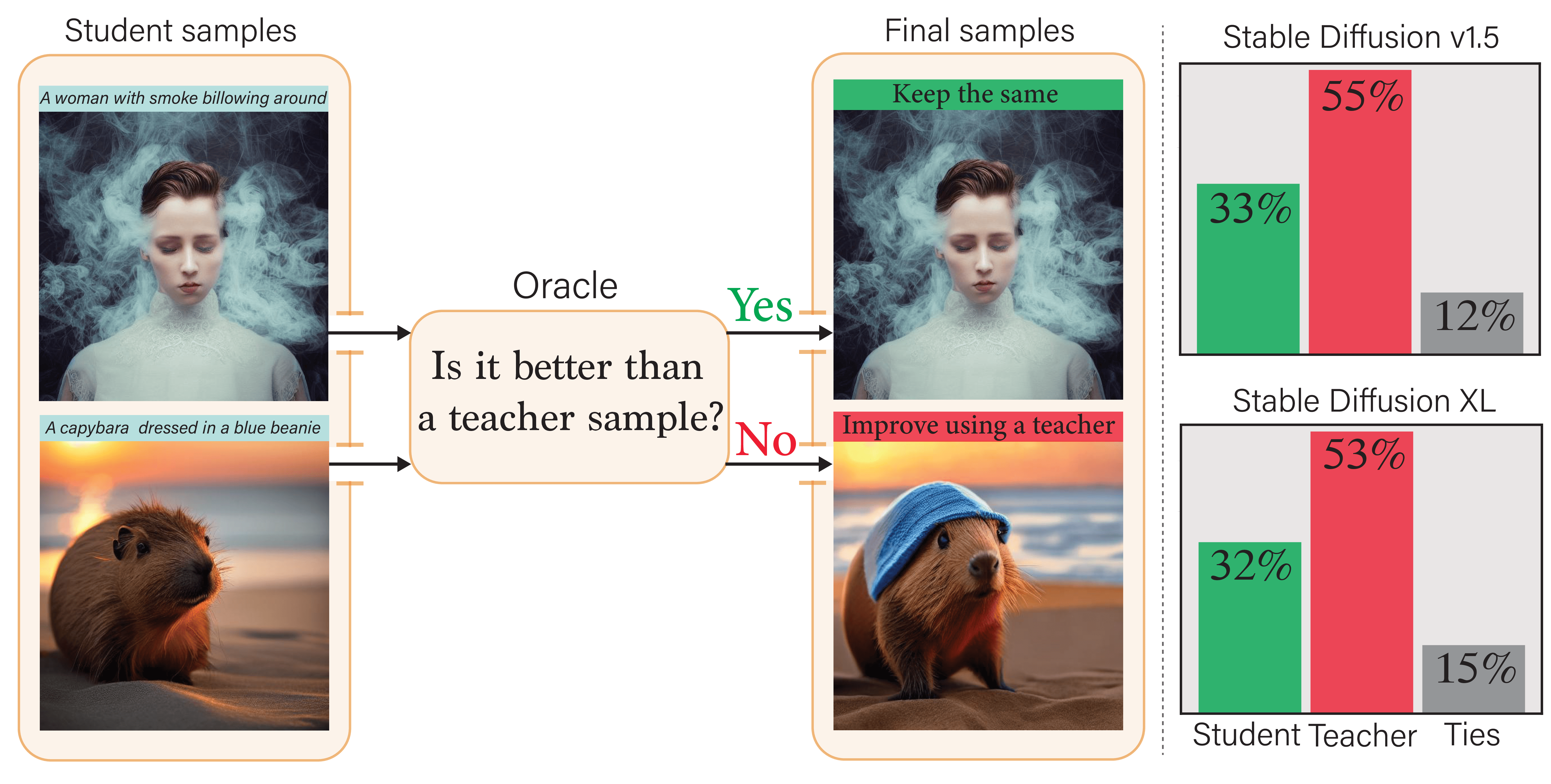}
    \vspace{-4mm}
    \captionof{figure}{
        \textit{Left:} Overview of the proposed approach. %a) The text-to-image student produces the initial sample; b) The oracle predicts if this sample is likely to be better than the teacher one generated for the same text prompt and noise sample; c) If yes, return the student sample. Otherwise, improve the sample using the teacher model.
        \textit{Right:} Side-by-side comparison of SDv$1.5$ and SDXL with their few-step distilled versions. The distilled models surpass the original ones in a noticeable number of samples for the same text prompts and initial noise. 
    }
    \vspace{0mm}
    \label{fig:main}
\end{center}%
}]

% Submission
\iffalse
\input{sec/0_abstract}    
\input{sec/1_intro}
\input{sec/2_related}
\input{sec/3_method}
\input{sec/5_exps}
\input{sec/6_conclusion}
\newpage
{
    \small
    \bibliographystyle{ieeenat_fullname}
    \bibliography{main}
}
\input{sec/X_suppl}
\fi

% Arxiv
\input{sec/0_abstract}    
\input{sec/1_intro_camera}
\input{sec/2_realated_camery}
\input{sec/3_method_camera}
\input{sec/5_exps_camera}
\input{sec/6_conclusion}
\newpage
{
    \small
    \bibliographystyle{ieeenat_fullname}
    \bibliography{main}
}
\input{sec/X_suppl_camera}

\end{document}

%% file: preamble.tex
\usepackage[dvipsnames,table]{xcolor}

\usepackage{url}
\usepackage[utf8]{inputenc} % allow utf-8 input
\usepackage[T1]{fontenc}    % use 8-bit T1 fonts
\usepackage{url}            % simple URL typesetting
\usepackage{booktabs}       % professional-quality tables
\usepackage{amsfonts}       % blackboard math symbols
\usepackage{nicefrac}       % compact symbols for 1/2, etc.
\usepackage{microtype}      % microtypography
\usepackage{lipsum}
\usepackage{fancyhdr}       % header
\usepackage{graphicx}       % graphics
\graphicspath{{media/}}     % organize your images and other figures under media/ folder
\usepackage{amsmath}
\usepackage{anyfontsize}
\usepackage[utf8]{inputenc} % allow utf-8 input
\usepackage[T1]{fontenc}    % use 8-bit T1 fonts
\usepackage{booktabs}       % professional-quality tables
\usepackage{amsfonts}       % blackboard math symbols
\usepackage{nicefrac}       % compact symbols for 1/2, etc.
\usepackage{microtype}      % microtypography
\usepackage{xcolor}         % colors
\usepackage{epsfig}
\usepackage{graphicx}
\usepackage{amsmath}
\usepackage{amssymb}
\usepackage{array}
\usepackage{caption}
\usepackage{subcaption}
\usepackage{xcolor}
\usepackage{tabularx}
\usepackage{booktabs}
\usepackage{multirow}
\usepackage{graphicx}
\usepackage{setspace}
\usepackage{amsmath,amssymb}
\usepackage{float}

\usepackage{amsthm}

\usepackage{derivative}
\usepackage{mathtools}
\usepackage{lipsum}
\usepackage{xcolor}
\usepackage{color}
\usepackage{wrapfig}
\usepackage[linesnumbered,ruled,vlined]
{algorithm2e}
\SetCommentSty{mycommfont}
\usepackage{tcolorbox}

\SetKwInput{KwInput}{Input}                % Set the Input
\SetKwInput{KwOutput}{Output}              % set the Output

\usepackage{amsthm}
\usepackage{amsmath}
\usepackage{amsfonts}
\usepackage{soul}
\usepackage{algorithm2e}

\newcommand{\fig}[1]{Figure~\ref{fig:#1}}

\newcommand{\tab}[1]{Table~\ref{tab:#1}}

%% file: sec/0_abstract.tex
\begin{abstract}

Knowledge distillation methods have recently shown to be a promising direction to speedup the synthesis of large-scale diffusion models by requiring only a few inference steps.
%These methods learn the student model that approximates the teacher distribution with only a few inference steps.
While several powerful distillation methods were recently proposed, the overall quality of student samples is typically lower compared to the teacher ones, which hinders their practical usage. 
In this work, we investigate the relative quality of samples produced by the teacher text-to-image diffusion model and its distilled student version. 
As our main empirical finding, we discover that a noticeable portion of student samples exhibit superior fidelity compared to the teacher ones, despite the ``approximate'' nature of the student.
Based on this finding, we propose an adaptive collaboration between student and teacher diffusion models for effective text-to-image synthesis. 
Specifically, the distilled model produces an initial image sample, and then an oracle decides whether it needs further improvements with the teacher model.
Extensive experiments demonstrate that the designed pipeline surpasses state-of-the-art text-to-image alternatives for various inference budgets in terms of human preference. 
Furthermore, the proposed approach can be naturally used in popular applications such as text-guided image editing and controllable generation.
\end{abstract}

%% file: sec/1_intro_camera.tex
\section{Introduction}
\label{sec:intro}

Large-scale diffusion probabilistic models (DPMs) have recently shown remarkable success in text-conditional image generation~\cite{podell2023sdxl, rombach2022high, saharia2022photorealistic, nichol2021glide} that aims to produce high quality images closely aligned with the user-specified text prompts.   
However, DPMs pose sequential synthesis leading to high inference costs opposed to feed-forward alternatives, e.g., GANs, that provide  decent text-to-image generation results for a single forward pass~\cite{Sauer2023stylegant, kang2023gigagan}. 

There are two major research directions mitigating the sequential inference problem of state-of-the-art diffusion models. 
One of them considers the inference process as a solution of a probability flow ODE and designs efficient and accurate solvers~\cite{song2020score, karras2022elucidating,lu2022dpm,luu2022dpm, zhao2023unipc} reducing the number of inference steps down to ${\sim}10$ without drastic loss in image quality.
Another direction represents a family of knowledge distillation approaches~\cite{salimans2022progressive, song2023consistency, meng2023distillation, liu2022flow, liu2023instaflow, luo2023latent, gu2023boot, sauer2023adversarial} that learn the student model to simulate the teacher distribution requiring only $1{-}4$ inference steps.
Recently, distilled text-to-image models have made a significant step forward~\cite{meng2023distillation, liu2023instaflow, luo2023latent, sauer2023adversarial}.
However, they still struggle to achieve the teacher performance either in terms of image fidelity and textual alignment~\cite{meng2023distillation, liu2023instaflow, luo2023latent} or distribution diversity~\cite{sauer2023adversarial}.
Nevertheless, we hypothesize that text-to-image students may already have qualitative merits over their teachers.
If so, perhaps it would be more beneficial to consider a teacher-student collaboration rather than focusing on replacing the teacher model entirely.

% \dmitry{reformulate: Recently, the distillation methods have made a significant step forward~\cite{meng2023distillation, liu2023instaflow, luo2023latent, sauer2023adversarial} but still struggle to reach the teacher performance, especially on highly challenging and diverse distributions that are currently standard for text-conditional generation~\cite{schuhmann2022laion}. 
% Nevertheless, it is unclear if text-to-image students already have qualitative advantages over their teachers.
% If so, perhaps it would be more beneficial to consider a student-teacher collaboration instead of aiming to replace the teacher model entirely.}

In this paper, we take a closer look at images produced by distilled text-conditional diffusion models and observe that the student can generate some samples even better than the teacher. 
Surprisingly, the number of such samples is significant and sometimes reaches up to half of the empirical distribution.
Based on this observation, we design an adaptive collaborative pipeline that leverages the superiority of student samples and outperforms both individual models alone for various inference budgets.
Specifically, the student model first generates an initial image sample given a text prompt, and then an ``oracle'' decides if this sample should be updated using the teacher model at extra compute.
The similar idea has recently demonstrated its effectiveness for large language models (LLMs)~\cite{chen2023frugalgpt} and we show that it can be naturally applied to text-conditional diffusion models as well.
Our approach is schematically presented in \fig{main}.
To summarize, our paper presents the following contributions:
%aims to determine if this sample is likely to be better than the teacher one generated for the same text prompt and initial noise.

\begin{itemize}
    \item We reveal that the distilled student DPMs can outperform the corresponding teacher DPMs for a noticeable number of generated samples.
    We demonstrate that most of the superior samples correspond to the cases when the student model significantly diverges from the teacher.
    \item 
    Based on the finding above, we develop an adaptive teacher-student collaborative approach for effective text-to-image synthesis. The method not only reduces the average inference costs but also improves the generative quality by exploiting the superior student samples.
    \item We provide an extensive human preference study illustrating the advantages of our approach for text-to-image generation. Moreover, we demonstrate that our pipeline can readily improve the performance of popular text-guided image editing and controllable generation tasks.
\end{itemize}

%% file: sec/2_realated_camery.tex
\section{Related work}

\textbf{Diffusion Probabilistic Models} (DPMs)~\cite{ho2020denoising, song2019generative, song2020score} represent a class of generative models consisting of \textit{forward} and  \textit{reverse} processes.
The \textit{forward} process $\{\boldsymbol{x}_t\}_{[0, T]}$ transforms real data $\boldsymbol{x}_{0} \sim p_{\text{data}}(\boldsymbol{x}_{0})$ into the noisy samples $\boldsymbol{x}_{t}$ using the transition kernels
$\mathcal{N}\left(\boldsymbol{x}_{t} \ | \ \sqrt{1 - \sigma_{t}} \boldsymbol{x}_{0}, \sigma_{t} \mathbf{I} \right)
    \label{eq:forward_transition}$
specifying $\sigma_t$ according to the selected \textit{noise schedule}.
%There are two prevalent special cases of diffusion processes: \textit{variance-preserving} (VP) and \textit{variance-exploding} (VE)\cite{scoresde}, where $\alpha_t = \sqrt{1 - \sigma^2_t}$ and $\alpha^2_t = 1$, respectively.
%The popular open-source text-conditional models, e.g., Stable Diffusion~\cite{}, usually operate with VP processes. 

The \textit{reverse} diffusion process generates new data points by gradually denoising samples from a simple (usually standard normal) distribution.
This process can be formulated as a \textit{probabilistic-flow ODE} (PF-ODE)~\cite{song2020denoising, song2020score}, where 
the only unknown component is a \textit{score function}, which is approximated with a neural network.
The ODE perspective of the reverse process fosters designing a wide range of the specialized solvers~\cite{lu2022dpm, zhao2023unipc, luu2022dpm, song2020denoising, karras2022elucidating, zhang2022fast, jolicoeur2021gotta} for efficient and accurate sampling.
However, for text-to-image generation, one still needs ${\sim}25$ and more steps for the top performance. 
% \begin{equation}
%     \begin{gathered}
%         \mathrm{d}\boldsymbol{x}= \left[\boldsymbol{f}(t)\boldsymbol{x}_t - g^2(t) \nabla_{\boldsymbol{x}}\log p_{t}(\boldsymbol{x}_t)\right]\mathrm{d}t,
%     \end{gathered} 
%     \label{eq:reverse_ode}
% \end{equation}
% where $f(t){=}\frac{\mathrm{d}\log \alpha_t}{\mathrm{d}t}$ and $g^2(t){=} \frac{\mathrm{d} \sigma^2_t}{\mathrm{d}t} - 2 \frac{\mathrm{d}\log \alpha_t}{\mathrm{d}t}\sigma^2_t$~\cite{vdp}.
%%%%%%%%%%%%%%%%%%%%%%%%%%%%%%%%%%%%%%%%%%%%%%%%%%%%%%%%%%%%%%%%%%%
% TODO: move it to analysis/experiments (setup description)
% In our work, we use DDIM~\cite{} and DPM-Solver~\cite{dpm2} as sampling methods. 
% DPM-Solver is described in detail in \ref{app:fast_solvers}.
%%%%%%%%%%%%%%%%%%%%%%%%%%%%%%%%%%%%%%%%%%%%%%%%%%%%%%%%%%%%%%%%%%%

\textbf{Text-conditional diffusion models} %inject the textual information into a neural network via textual embedding extracted using a text-encoder, e.g., CLIP~\cite{radford2021learning}. 
%To attain a trade-off between sample quality and diversity, the classifier-free~\cite{ho2022classifier} or classifier guidance~\cite{dhariwal2021diffusion} can be applied. 
can be largely grouped into \textit{cascaded} and \textit{latent} diffusion models. 
The cascaded models~\cite{saharia2022photorealistic, nichol2021glide} generate a sample in several stages using separate diffusion models for different image resolutions.
The latent diffusion models~\cite{podell2023sdxl, rombach2022high, ramesh2021zero} first generate a low-resolution latent variable in the VAE~\cite{kingma2013auto} space and then apply its feedforward decoder to map the latent sample to the high-resolution pixel space. 
Thus, the latent diffusion models have significantly more efficient inference thanks to a single forward pass for the upscaling step.  

\begin{figure*}[ht!]
    \centering
    \includegraphics[width=\linewidth]{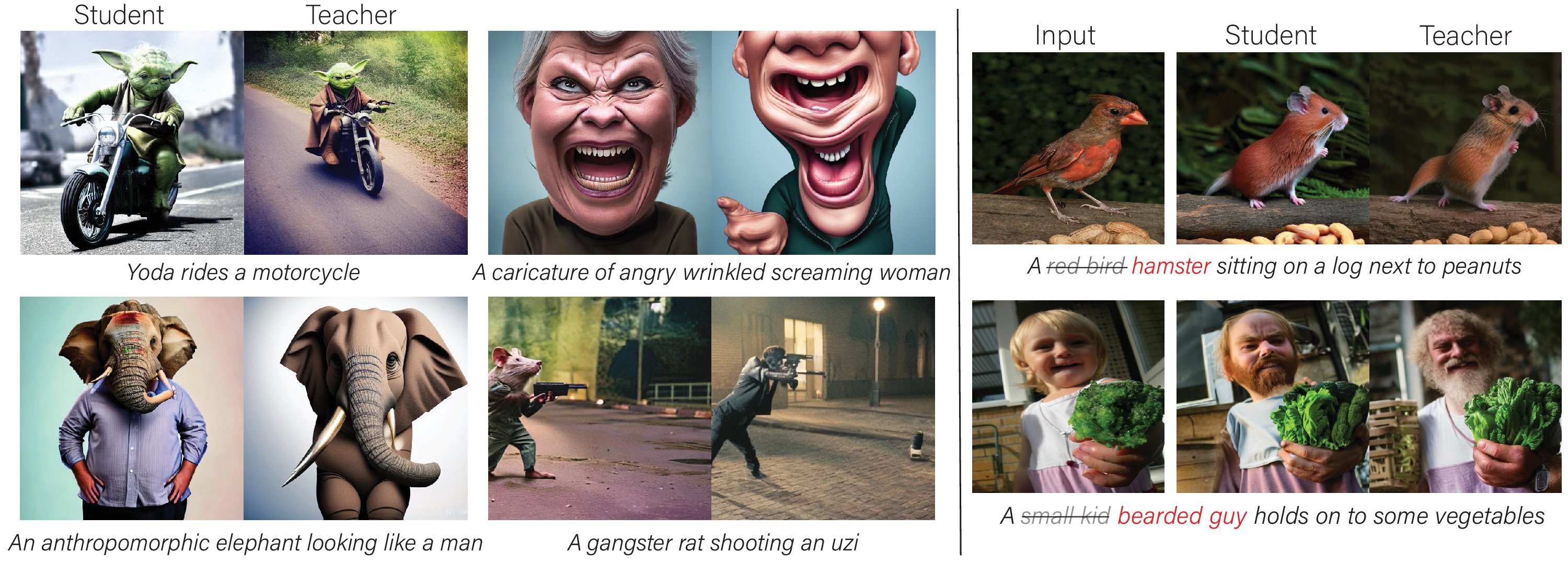}
    \vspace{-6mm}
    \captionof{figure}{\textbf{Student outperforms its teacher (SD1.5)}. \textit{Left:}  Text-conditional image synthesis. \textit{Right:} Text-guided image editing (SDEdit~\cite{meng2021sdedit}). The images within each pair are generated for the same initial noise sample.} 
    \label{fig:student_better_teacher}
\end{figure*}

Along with cascaded models, there are other works combining several diffusion models into a single pipeline.
Some methods propose to use distinct diffusion models at different time steps~\cite{balaji2022ediffi, liu2023oms, feng2023ernie, xue2023raphael}.
Others~\cite{podell2023sdxl, liu2023instaflow} consider an additional model to refine the samples produced with a base model. 
In contrast, our method relies on the connection between student and teacher models and adaptively improves only selected student samples to reduce the inference costs.

Text-to-image diffusion models have also succeeded in text-guided image editing and personalization~\cite{brooks2023instructpix2pix, meng2021sdedit, mokady2022null, kawar2023imagic, ruiz2022dreambooth, gal2022textual}. 
Moreover, some methods allow controllable generation via conditioning on additional inputs, e.g., canny-edges, semantic masks, sketches~\cite{zhang2023adding, voynov2023sketch}.
Our experiments show that the proposed pipeline is well-suited to these techniques. 

\textbf{Distillation of diffusion models} is another pivotal direction for efficient diffusion inference~\cite{song2023consistency, song2023improved, salimans2022progressive, meng2023distillation, berthelot2023tract, liu2023instaflow, sauer2023adversarial}. 
The primary goal is to adapt the diffusion model parameters to represent the teacher image distribution for $1{-}4$ steps. 
Recently, consistency distillation (CD)~\cite{song2023consistency} have demonstrated promising results on both classical benchmarks~\cite{kim2023consistency, song2023improved} and text-to-image generation~\cite{luo2023latent} but fall short of the teacher performance at the moment.
Concurrently, adversarial diffusion distillation~\cite{sauer2023adversarial} 
could outperform the SDXL-Base~\cite{podell2023sdxl} teacher for $4$  steps in terms of image quality and prompt alignment.
However, it significantly reduces the diversity of generated samples, likely due to the adversarial training~\cite{goodfellow2020generative} and mode-seeking distillation technique~\cite{poole2023dreamfusion}.
Therefore, it is still an open question if a few-step distilled model can perfectly approximate the diffusion model on highly challenging and diverse distributions that are currently standard for text-conditional generation~\cite{schuhmann2022laion}.  

% Nevertheless, the distilled models still fall short of the teacher performance for image generation~\cite{song2023consistency}.

%% file: sec/3_method_camera.tex
\section{Toward a unified teacher-student framework} \label{main_analysis}

Opposed to the purpose of replacing the expensive text-to-image diffusion models by more effective few-step alternatives, the present work suggests considering the distilled text-to-image models as a firm companion in a teacher-student collaboration.

In this section, we first explore the advantages of the distilled text-to-image models and then unleash their potential in a highly effective generative pipeline comprising the student and teacher models. 

\subsection{Delving deeper into the student performance} 
We start with a side-by-side comparison of the student and teacher text-to-image diffusion models. 
Here, we focus on Stable Diffusion v1.5\footnote{https://huggingface.co/runwayml/stable-diffusion-v1-5} (SD$1.5$) as our main teacher model and distill it using consistency distillation~\cite{song2023consistency}.
The student details and sampling setup are presented in~\ref{app:student_details}.
The similar analysis for a few other distilled models is provided in~\ref{app:other_distill}.
% We also investigate the recent latent consistency model~\cite{luo2023latent} distilled from Dreamshaper v7\footnote{https://huggingface.co/Lykon/dreamshaper-7} in~\ref{app:analysis}.

\begin{figure}[ht!]
    \centering
    \vspace{-1mm}
\includegraphics[width=1.0\linewidth]{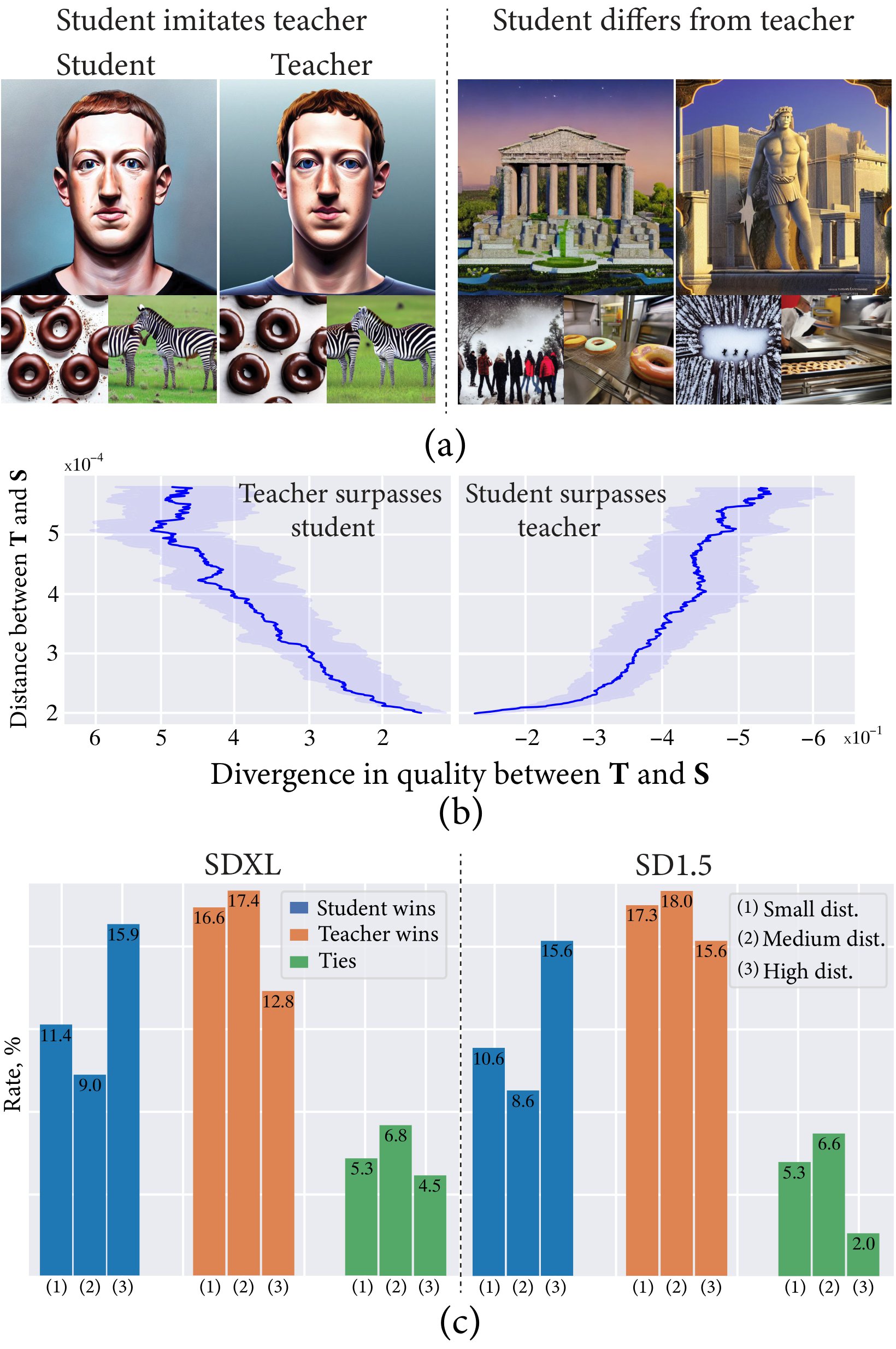}
    \vspace{-7mm}
    \caption{(a) Visual examples of similar (Left) and dissimilar (Right) teacher and student samples. 
    (b) Similarity between the student and teacher samples w.r.t. the difference in sample quality. 
    Highly distinct samples tend to be of different quality. 
    (c) Human vote distribution for different distance ranges between student and teacher samples. Most of the student wins are achieved when the student diverges from the teacher.
    %(*) The histograms represent individual distributions and are not comparable between each other.
    }
    \vspace{-1mm}
    \label{fig:dual_mode_beh}
\end{figure}
\begin{figure}[t!]
    \centering
    \vspace{2mm}
    \includegraphics[width=1.0\linewidth]{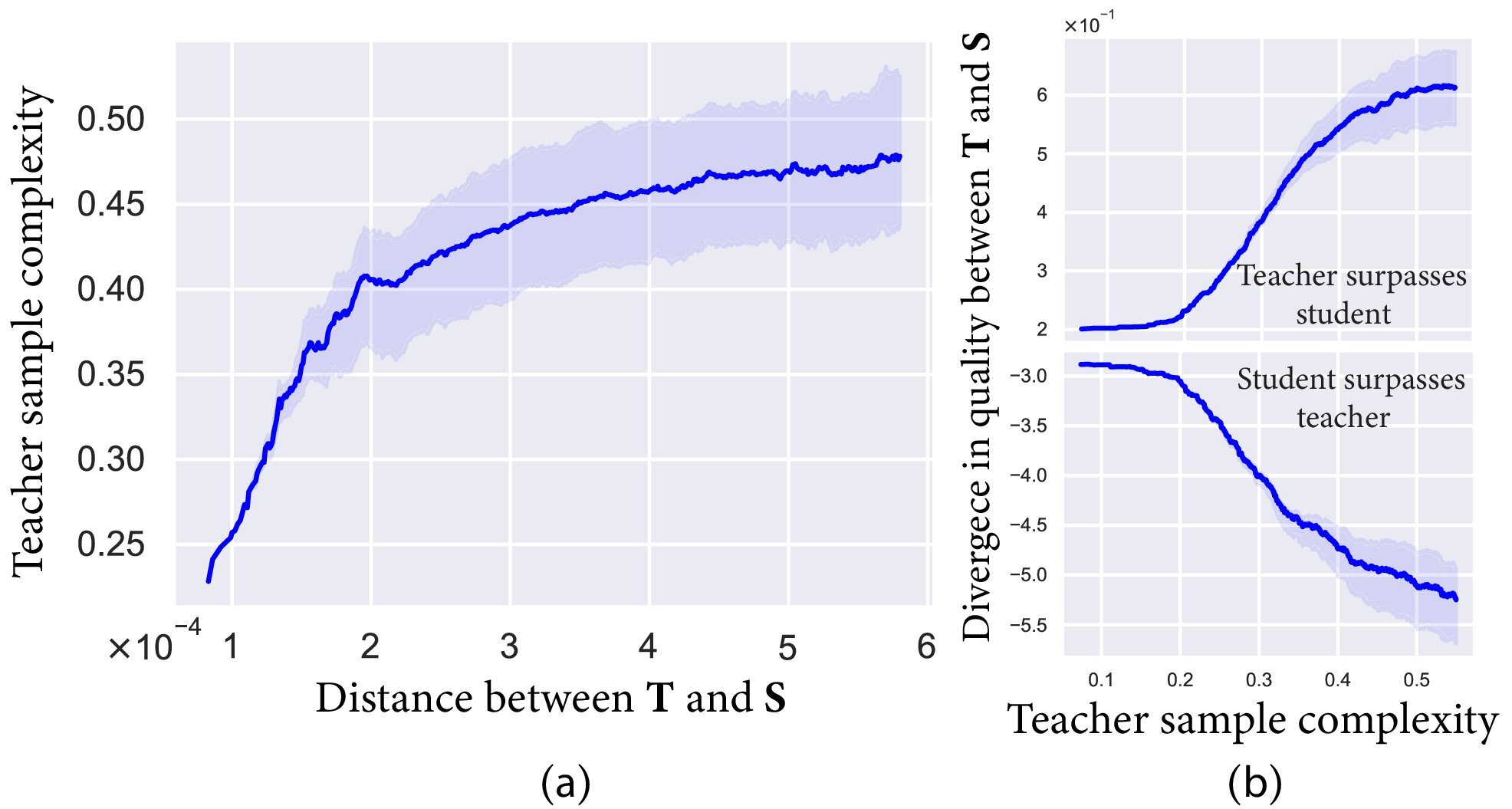}
    \vspace{-6mm}
    \caption{\textbf{Effect of image complexity}. 
    (a) More similar student and teacher samples corresponds to simpler images and vice versa. 
    (b) The student and teacher largely diverge in image quality on the complex teacher samples. }
    \label{fig:image_complexity}
\end{figure} 
\begin{figure}[t!]
    \centering
    \includegraphics[width=1.0\linewidth]{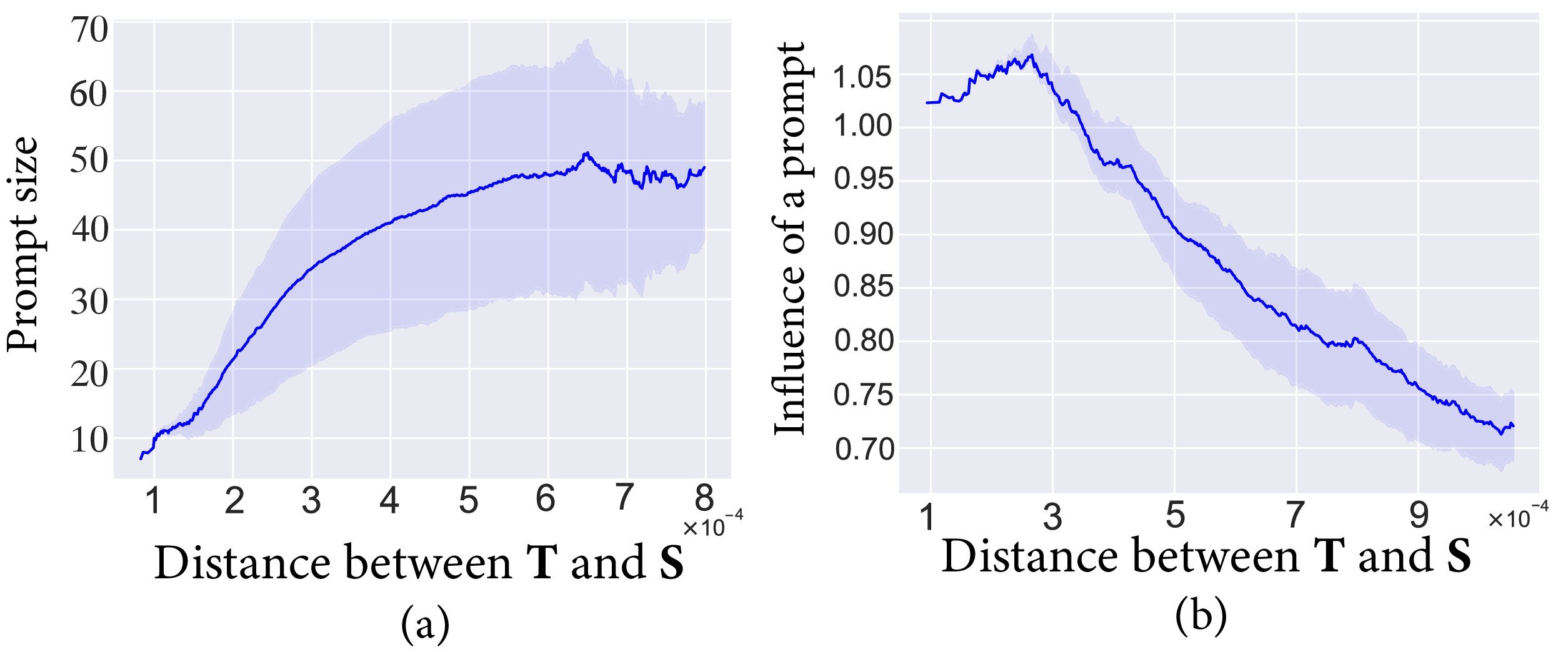}
    \vspace{-6mm}
    \caption{\textbf{Effect of text prompts}. 
    (a) Shorter prompts usually lead to more similar student and teacher samples.
    (b) The student and teacher tend to generate more similar images when the student relies heavily on the text prompt.}
    \label{fig:prompts}
\end{figure}

In \fig{main}~(Right), we provide the human votes for 600 random text prompts from COCO2014~\cite{lin2014microsoft} for SD$1.5$  and SDXL.
The images within each pair are generated for the same initial noise sample.
We observe that the students generally falls short of the teacher performance.
However, interestingly, despite the initial intention to mimic the teacher model, ${\sim}30\%$ student samples were preferred over the teacher ones.
A few visual examples in ~\fig{student_better_teacher} validate these results.
% More examples are in \ref{app:analysis}. 
Therefore, we can formulate our first observation:
\begin{tcolorbox}[colback=red!5!white,colframe=black!75!black]
The student can surpass its teacher in a substantial portion of image samples.\end{tcolorbox} 
Below, we develop a more profound insight into this phenomenon.
% For evaluation, we consider 5000 prompts from the COCO2014~\cite{} validation split and a random subset of 600 prompts for human evaluation.
% Note that we align the student and teacher samples by fixing the initial noise sample $x_T\sim N(0,I)$ within each prompt.
% Concurrently, \citet{lcm} have realised the latent consistency model distilled from DreamShaper v7~\cite{}.
% We provide the similar analysis for this model in \app{dreamshaper}.
% There is also publicly available single-step InstaFlow~\cite{} model but it fails to produce satisfactory samples in our experiments.
\\ \textbf{Student-teacher similarity.}
First, we evaluate the student's ability to imitate the teacher.
We compute pairwise distances between the student (S) and teacher (T) images generated for the same text prompts and initial noise.
As a distance measure, we use DreamSim~\cite{fu2023dreamsim} tuned to be aligned with the human perception judgments. 
For evaluation, we consider 5000 prompts from the COCO2014~\cite{lin2014microsoft} validation split.

Primarily, we observe that many student samples are highly distinct from the teacher ones. 
A few image pairs are presented in Figure~\ref{fig:dual_mode_beh}a. 
\fig{dual_mode_beh}c presents the human vote distribution for low ($0{-}20\%$), medium ($40{-}60\%$) and high ($80{-}100\%$) distance ranges. %that correspond to $0{-}20\%$, $40{-}60\%$ and $80{-}100\%$ of distances, respectively.
Interestingly, most of the student wins appear when its samples are highly different from the teacher ones. 
This brings us to our second observation:
\begin{tcolorbox}[colback=red!5!white,colframe=black!75!black]
The student wins are more likely where its samples significantly differ from the teacher ones.  \end{tcolorbox}

Also, we evaluate the relative gap in sample quality against the similarity between the teacher and student outputs.
To measure the quality of individual samples, we use ImageReward~\cite{xu2023imagereward}, which shows a positive correlation with human preferences in terms of image fidelity and prompt alignment. 
The divergence in quality is calculated as the difference between the ImageReward scores for student and teacher samples.
We observe that highly distinct samples likely have a significant difference in quality.
Importantly, this holds for both student failures and successes, as shown in \fig{dual_mode_beh}b.
Therefore, effectively detecting the positive student samples and improving the negative ones can potentially increase the generative performance. 
\\ \textbf{Image complexity.}
Then, we describe the connection of the similarity between student and teacher samples with the teacher image complexity. 
To estimate the latter, we use the ICNet model~\cite{feng2023ic9600} learned on a large-scale human annotated dataset.  
The results are presented in \fig{image_complexity}. 
We notice that larger distances between student and teacher outputs are more typical for complex teacher samples. 
In other words, the student mimics its teacher for plain images, e.g., close-up faces, while acting more as an independent model for more intricate ones. 
\fig{image_complexity}b confirms that significant changes in image quality are observed for more complex images.
\\ \textbf{Text prompts.}
Then, we analyse the connection of the student-teacher similarity with the prompt length.
\fig{prompts} demonstrates that shorter prompts typically lead to more similar teacher and student samples.
Here, the prompt length equals to the number of CLIP tokens. 
Intuitively, longer prompts are more likely to describe intricate scenes and object compositions than shorter ones.
Note that long prompts can also carry low textual informativeness and describe concepts of low complexity. 
We hypothesize that this causes high variance in \fig{prompts}a. 

\begin{figure}[t!]
    \centering
    \includegraphics[width=1.0\linewidth]{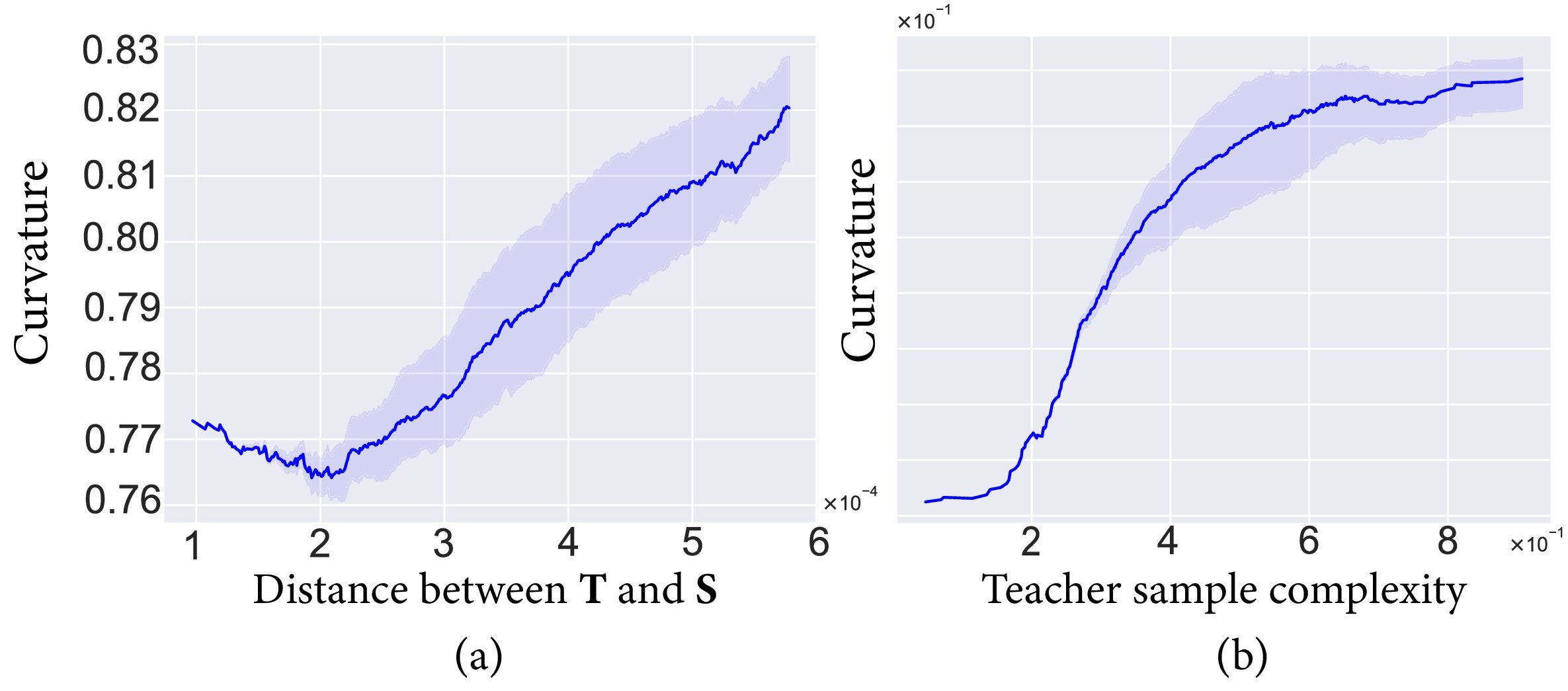}
    \vspace{-6mm}
    \caption{\textbf{Effect of teacher trajectory curvature}. 
    (a) The student samples resemble the teacher ones for less curved trajectories. 
    (b) Straighter trajectories usually correspond to plainer teacher images.}
    \label{fig:curvature}
\end{figure}

Also, we report the prompt influence on the student generation w.r.t. the student-teacher similarity in \fig{prompts}b. 
We estimate the prompt influence by aggregating student cross-attention maps. 
More details are in \ref{app:calc_details}.
The student tends to imitate the teacher if it relies heavily on the text prompt.
\\ \textbf{Trajectory curvature.}
Previously, it was shown to be beneficial to straighten the PF-ODE trajectory before distillation~\cite{liu2022flow,liu2023instaflow}. 
We investigate the effect of the trajectory curvature on the similarity between the teacher and student samples and its correlation with the teacher sample complexity. 
We estimate the trajectory curvatures following~\cite{chen2023geometric} and observe that straighter trajectories lead to more similar student and teacher samples (\fig{curvature}a). 
In addition, we show that the trajectory curvature correlates with the teacher sample complexity (\fig{curvature}b).

\begin{figure}[t!]
    \centering
    \includegraphics[width=1.0\linewidth]{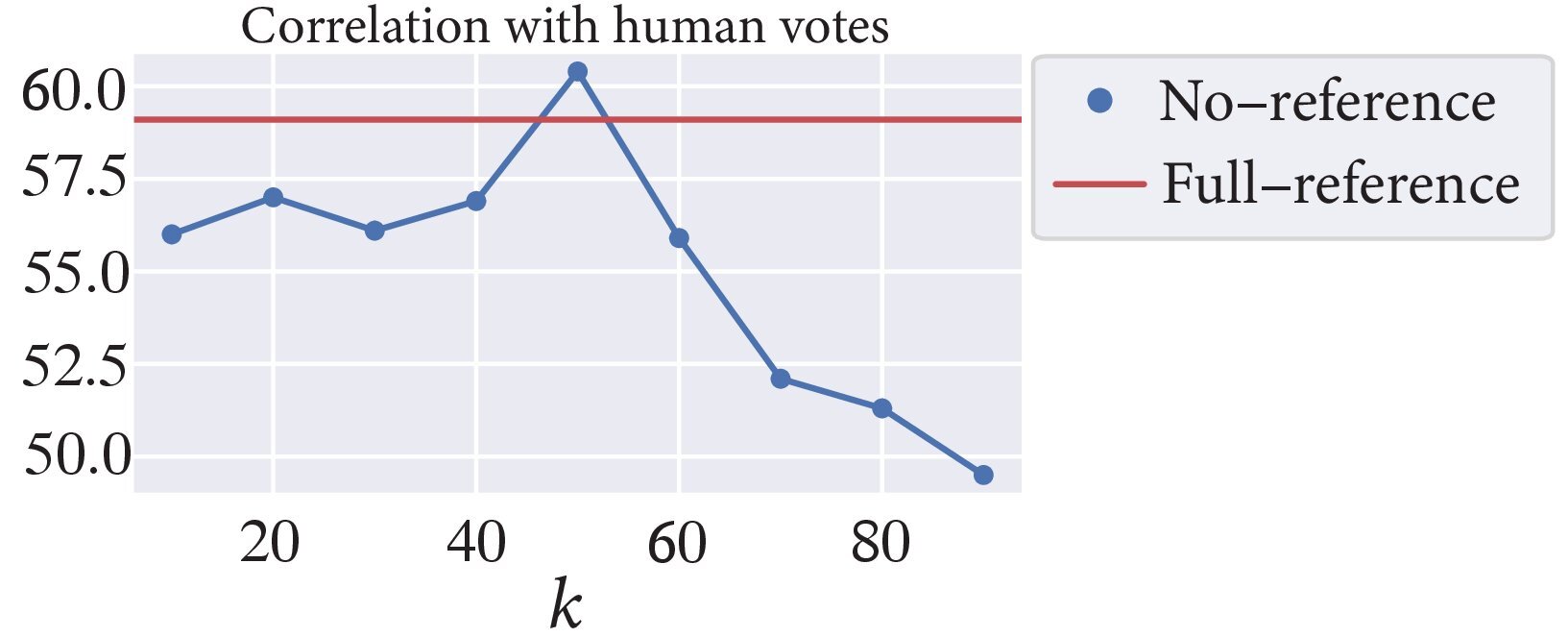}
    \caption{\textbf{Full-reference vs no-reference decision-making}. Usually, one can find a $k$-th percentile of the ImageReward scores providing the correlation with human votes similar to the full-reference comparisons but without observing the teacher samples.}
    %Usually, there is a percentile providing the same correlation as the full-reference estimate without seeing a teacher sample.}
    \label{fig:ref_no_ref}
\end{figure} 

To sum up, we conclude that the student largely diverges from the teacher on the samples that are challenging in different respects.
Interestingly, the superior student samples often occur in these cases.

% This behavior resembles deep ensembles~\cite{} where a bunch of independently trained models provide diverse predictions on the complex inputs and increase the overall performance.
% However, in our case, the distilled model initialized with the teacher weights and learnt to imitate the teacher still can produce sufficient diverse samples to significantly improve the teacher performance.
% We hypothesize that it is caused by the highly stochastic nature of the diffusion models and consistency distillation methods.

\subsection{Method}

In this section, we propose an adaptive collaborative approach consisting of three steps:
1) Generate a sample with the student model;
2) Decide if the sample needs further improvement;
3) If so, refine or regenerate the sample with the teacher model.

\textbf{Student generation step} produces an initial sample $\mathcal{X}^{\mathbf{S}}$ for a given context and noise.
This work considers consistency distillation~\cite{song2023consistency} as a primary distillation framework and uses multistep consistency sampling~\cite{song2023consistency} for generation. %\dmitry{what about ADD?}

\textbf{Adaptive step} leverages our finding that many student samples may exhibit superior quality.
Specifically, we seek an ``oracle'' that correctly detects superior student samples. 
For this role, we consider an individual sample quality estimator $\mathbf{E}$. 
In particular, we use the current state-of-the-art automated estimator, ImageReward (IR)~\cite{xu2023imagereward} that is learned to imitate human preferences for text-to-image generation.

Then, comparing the scores of the teacher and student samples, one can conclude which one is better.
However, in practice, we avoid expensive teacher inference to preserve the efficiency of our approach.
Therefore, a decision must be made having access only to the student sample $\mathcal{X}^{\mathbf{S}}$. 
To address this problem, we introduce a cut-off threshold $\tau$ which is a $k$-th percentile of the IR score tuned on a hold-out subset of student samples. 
The details on the threshold tuning are described in~\ref{app:method}.
During inference, the IR score is calculated only for $\mathcal{X}^{\mathbf{S}}$.
If it exceeds the threshold $\tau$, we accept the sample and avoid further teacher involvement. 
Interestingly, we observe that it is often possible to reproduce the accuracy of the full-reference estimation by varying $\tau$ (see \fig{ref_no_ref}). 
Also, note that IR calculation costs are negligible compared to a single diffusion step, see \ref{app:ir_inference_costs}.

\textbf{Improvement step} engages the teacher to improve the quality of the rejected student samples.
We consider two teacher involvement strategies: \textit{regeneration} and \textit{refinement}. 
The former simply applies the teacher model to produce a new sample from scratch for the same text prompt and noise.
The refinement is inspired by the recent work~\cite{podell2023sdxl}. 
Specifically, $\mathcal{X}^{\mathbf{S}}$ is first corrupted with a Gaussian noise controlled by the rollback value $\sigma \in [0, 1]$. 
Higher $\sigma$ leads to more pronounced changes. 
We vary $\sigma$ between $0.3$ and $0.75$ in our experiments.
Then, the teacher starts sampling from the corrupted sample following the original noise schedule and using an arbitrary solver, e.g., DPM-Solver~\cite{lu2022dpm}.
Note that refinement requires significantly fewer steps to produce the final sample than generation from scratch. 
Intuitively, the refinement strategy aims to fix the defects of the student sample.
At the same time, the regeneration strategy may be useful if $\mathcal{X}^{\mathbf{S}}$ is poorly aligned with the text prompt in general.
Our experiments below confirm this intuition.

%% file: sec/5_exps_camera.tex
\begin{figure*}[t!]
    \centering
    \includegraphics[width=0.99\linewidth]{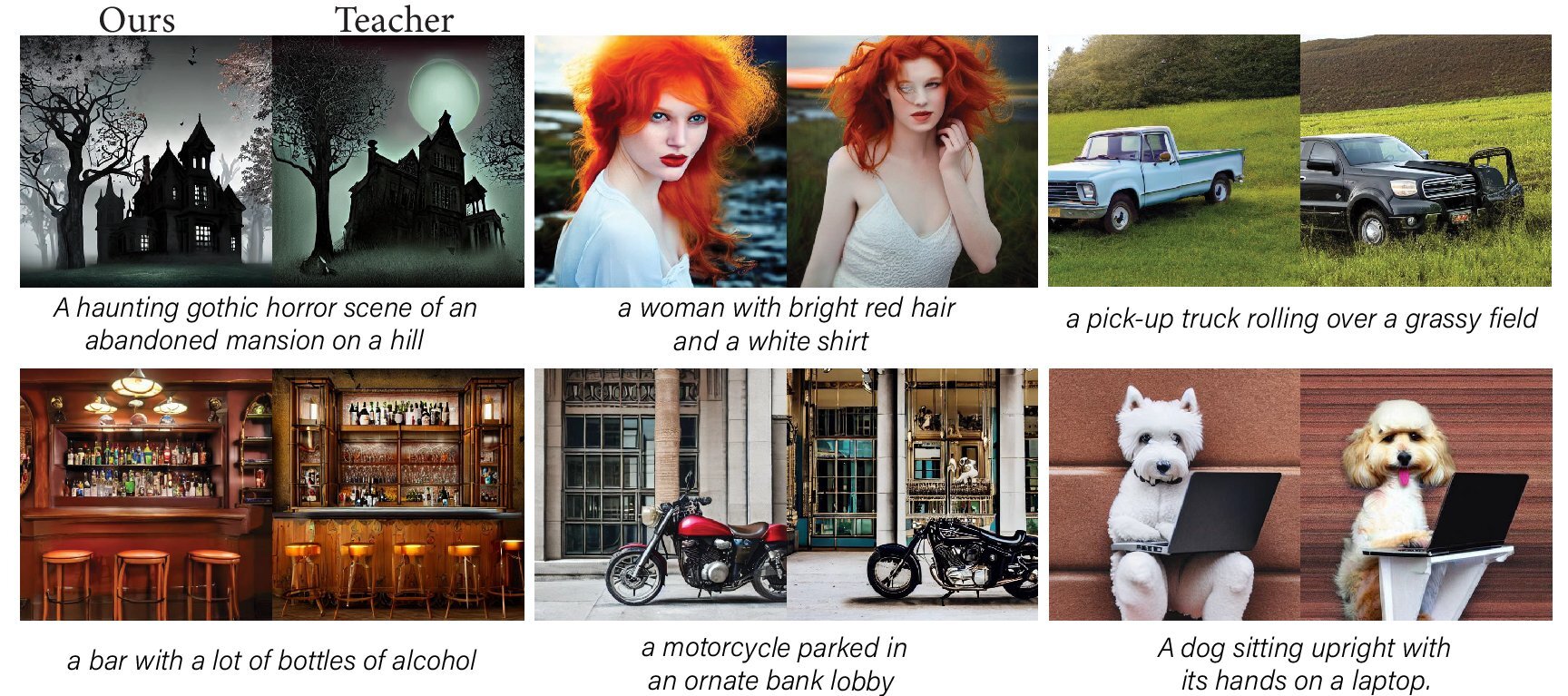}
    \vspace{-2mm}
    \captionof{figure}{Qualitative comparison of our adaptive refinement approach to the SD1.5 teacher model.} 
    \label{fig:ours_teacher_generation}
\end{figure*}

\begin{figure*}[t!]
    \centering
    \includegraphics[width=0.94\linewidth]{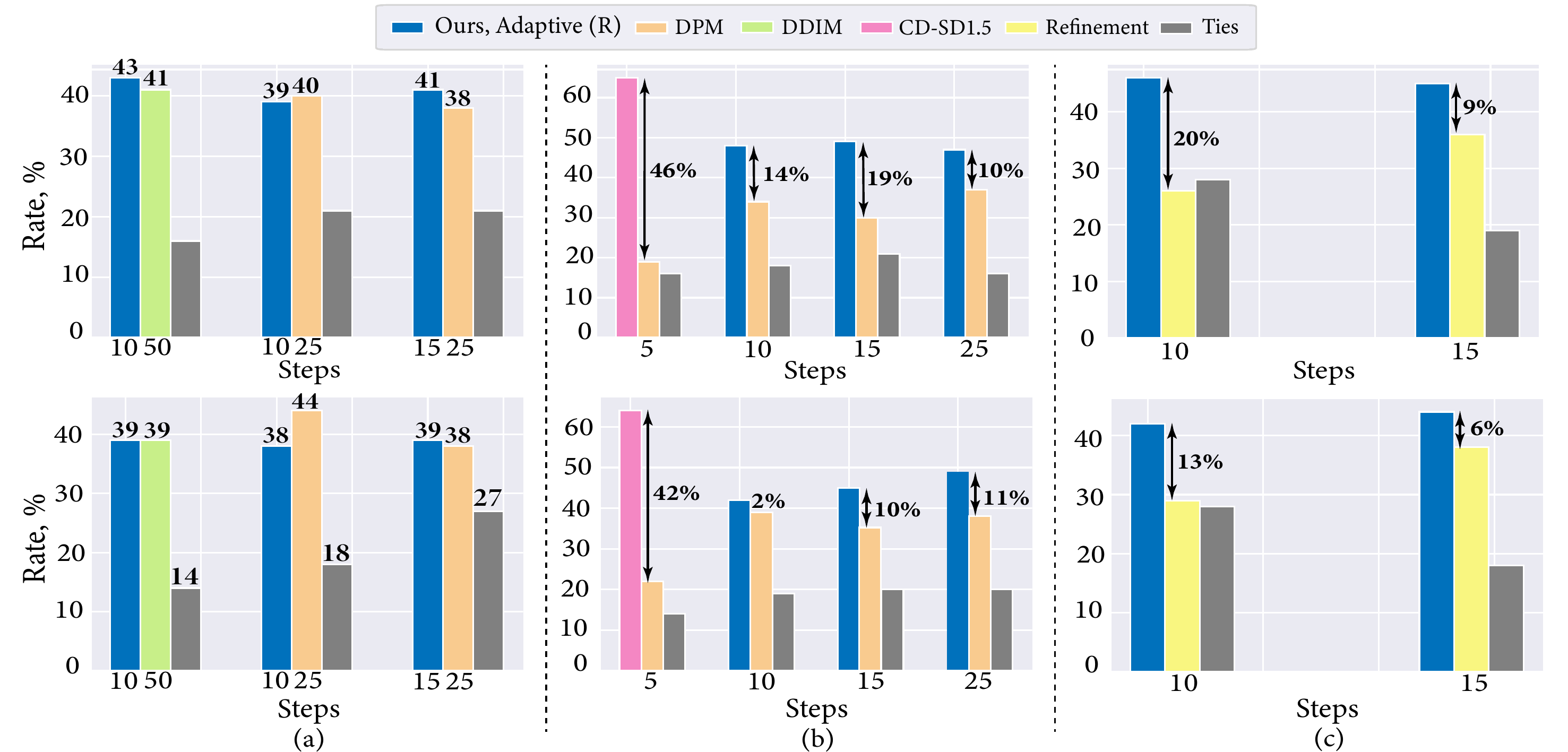}
    \vspace{-3mm}
    \captionof{figure}{\textbf{User preference study (SD1.5)}. 
    (a) Comparison of our approach to the top-performing teacher configurations.
    (b) Comparison to the teacher model with DPM-Solver for the same average number of steps.
    (c) Comparison to the refinement strategy without the adaptive step for the same average number of steps.
    \textit{Top row:} LAION-Aesthetic text prompts. 
    \textit{Bottom row:} COCO2014 text prompts.
    For our adaptive approach, we use the refinement strategy (R).
    } 
    \label{fig:users_study}
\end{figure*}

\section{Experiments}
We evaluate our approach for text-to-image synthesis, text-guided image editing and controllable generation. 
The results confirm that the proposed adaptive approach can outperform the baselines for various inference budgets. 

\subsection{Text-guided image synthesis} \label{subsec:image_synth}
In most experiments, we use Stable Diffusion v1.5 (SD1.5) as a teacher model and set the classifier-free guidance scale to $8$.
% All generated images have $512{\times}512$ resolution.
To obtain a student model, we implement consistency distillation (CD) for latent diffusion models and distill SD1.5 on the 80M subset of LAION2B~\cite{schuhmann2022laion}. 
The resulting model demonstrates decent performance for $5$ steps of multistep consistency sampling with the guidance scale $8$.
\\ \textbf{Metrics}. 
We first consider FID~\cite{heusel2017fid}, CLIP score~\cite{radford2021learning} and ImageReward~\cite{xu2023imagereward} as automatic metrics. 
ImageReward is selected due to a higher correlation with human preferences compared to FID and CLIP scores.
OpenCLIP ViT-bigG~\cite{cherti2022reproducible} is used for CLIP score calculation. 
For evaluation, we use $5000$ text prompts from the COCO2014 validation set~\cite{lin2014microsoft}. 

\begin{figure*}[t!]
    \centering
    \vspace{-1mm}
    \includegraphics[width=0.97\linewidth]{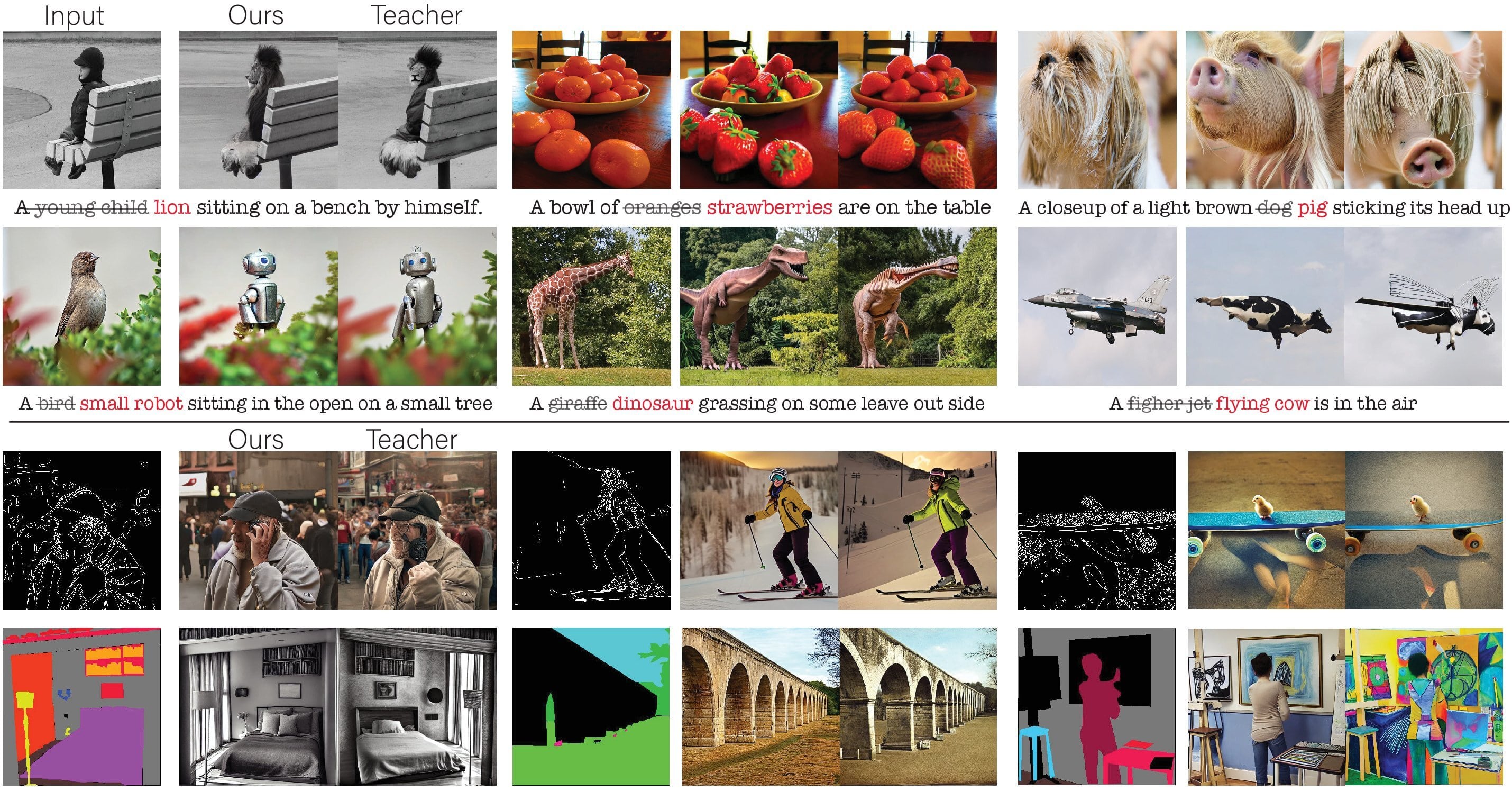}
    \vspace{-3mm}
    \caption{Visual examples produced with our approach and the top-performing teacher (SD1.5) configuration. \textit{Top:} Text-guided image editing with SDEdit~\cite{meng2021sdedit}. \textit{Bottom:} Controllable image generation with Canny edges and semantic segmentation masks using ControlNet~\cite{zhang2023adding}.} 
    \label{fig:ours_teacher}
\end{figure*}

Also, we evaluate user preferences using side-by-side comparisons conducted by professional assessors.
We select $600$ random text prompts from the COCO2014 validation set and $600$ from LAION-Aesthetics.
More details on the human evaluation pipeline are provided in~\ref{app:human_eval}. 
\\ \textbf{Configuration}. 
For our adaptive approach, we consider both refinement (R) and regeneration (G) strategies using a second order multistep DPM solver~\cite{luu2022dpm} and vary the number of sampling steps depending on the average inference budget. %from $5$ to $45$
As a sample estimator $\mathbf{E}$, we consider ImageReward, except for the CLIP score evaluation. 
For each inference budget, we tune the hyperparameters $\sigma$ and $\tau$ on the hold-out prompt set. 
The exact values are provided in~\ref{app:sd15_setup}.
\\ \textbf{Baselines}. 
We consider the teacher performance as our main baseline and use DDIM~\cite{song2020denoising} for 50 steps and a second order multistep DPM solver~\cite{luu2022dpm} for lower steps.  
In addition, we compare to the refining strategy on top of all student samples, without the adaptive step.
This baseline is inspired by the recent results~\cite{podell2023sdxl} demonstrating the advantages of the refinement stage itself.
Also, we provide the comparison with Restart Sampling~\cite{xu2023restart}.
\begin{figure}[t!]
    \centering
    \includegraphics[width=1.0\linewidth]{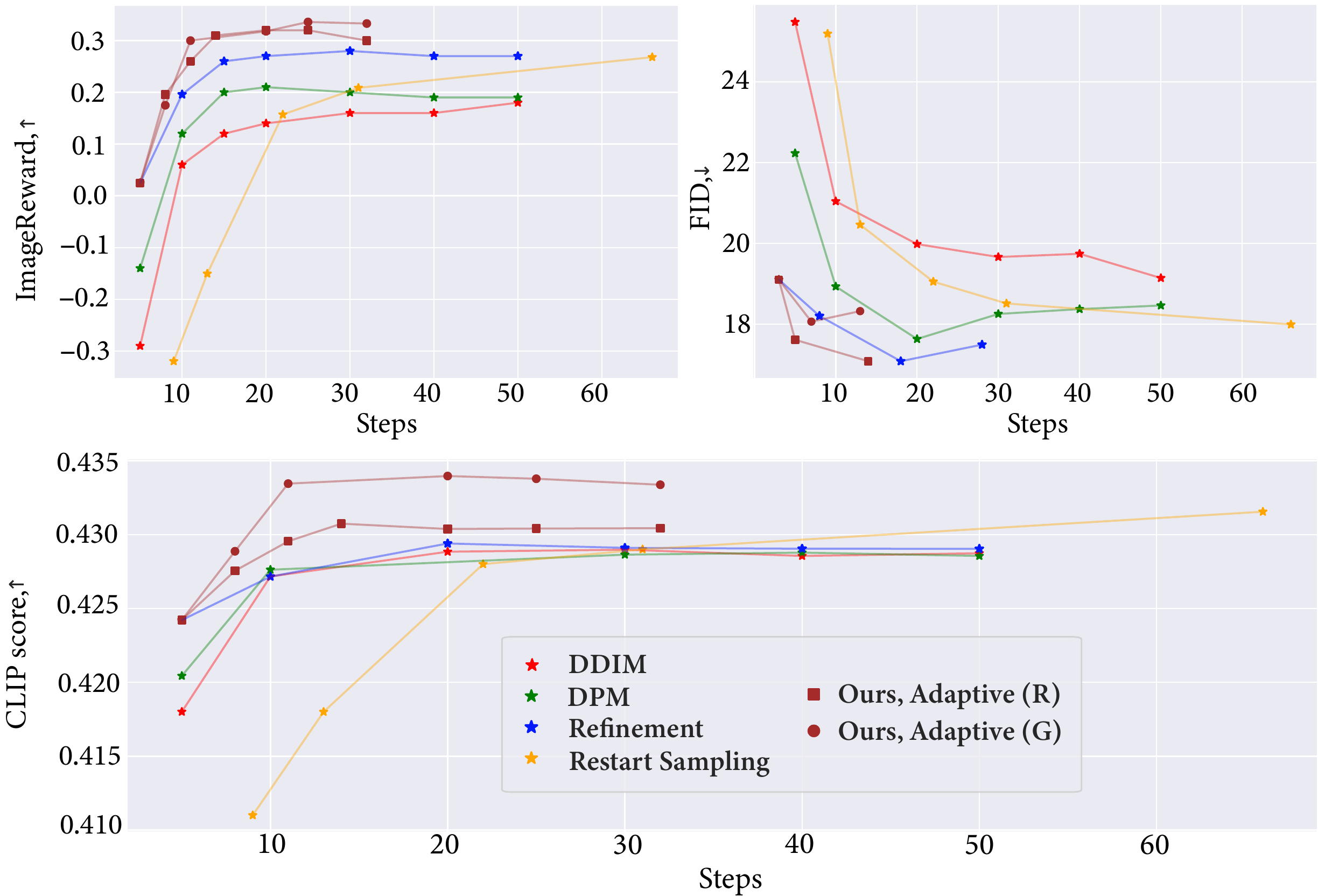}
    \vspace{-6mm}
    \caption{{\textbf{Automated evaluation (SD1.5)}. Comparison of the FID, CLIP and ImageReward scores for different number of sampling steps on 5K text prompts from COCO2014. 
    The proposed collaborative approach outperforms all the baselines.
    The adaptive pipeline with the regeneration strategy (G) demonstrates higher textual alignment (CLIP score), while the refinement strategy (R) improves the image fidelity (FID).
    }}
    \label{fig:automatic_t2i}
\end{figure} 
\begin{figure}[t!]
    \centering
    \includegraphics[width=1.0\linewidth]{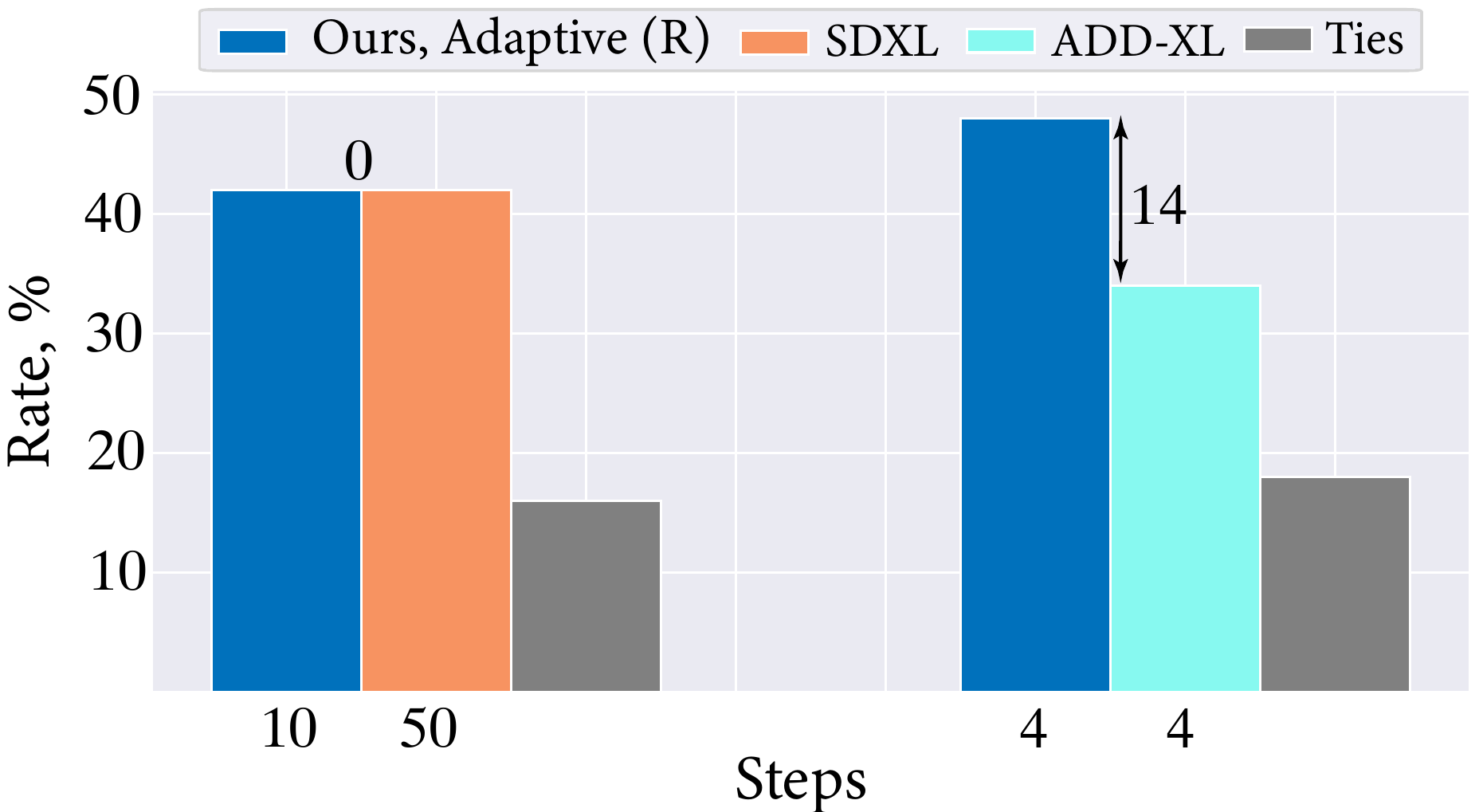}
    \vspace{-6mm}
    \caption{
        \textbf{User preference study (SDXL).} 
        \textit{Left:} Comparison of the adaptive approach with CD-SDXL to the top-performing teacher setup. 
        \textit{Right:} Comparison of the adaptive approach with ADD-XL to ADD-XL for the same average number of steps.
    }
    \label{fig:sdxl_res}
    \vspace{-3mm}
\end{figure} 
%\begin{figure}[t!]
%    \centering
%    \includegraphics[width=1.0\linewidth]{images/dynamic_acc.pdf}
%    \vspace{-6mm}
%    \caption{User preferences for different accuracy levels of the no-reference decision-making procedure.
    %the automated sample estimator. 
%    \textit{Current state} represents the results using ImageReward. 
%    The results for higher accuracy rates demonstrate the %future gains if the oracle performance improves.}
%    \label{fig:perspectives}
%\end{figure} 
\\ \textbf{Results}. 
The quantitative and qualitative results are presented in Figures \ref{fig:users_study}, \ref{fig:automatic_t2i} and Figure \ref{fig:ours_teacher}, respectively. 
According to the automatic metrics, our approach outperforms all the baselines. 
Specifically, in terms of CLIP scores, the adaptive regeneration strategy demonstrates superior performance compared to the refining-based counterpart. 
On the other hand, the adaptive refining strategy is preferable in terms of FID scores. 
We assume that the refinement strategy essentially improves the image fidelity and does not significantly alter the textual alignment due to the relatively small rollback values.  
In terms of ImageReward, both adaptive strategies perform equally. 

In the human evaluation, we consider two nominations: i) \textit{acceleration}, where our approach aims to reach the performance of SD1.5 using $50$ DDIM steps or $25$ DPM steps; ii) \textit{quality improvement}, where the adaptive method is compared to the baselines for the same average number of steps. 
The results for the acceleration nomination are presented in Figure \ref{fig:users_study}a.
The proposed method achieves the teacher performance for $5{\times}$ and up to $2.5{\times}$ fewer steps compared to DDIM$50$ and DPM$25$, respectively.
The results for the second nomination (Figure \ref{fig:users_study}b,c) confirm that our approach consistently surpasses alternatives using the same number of steps on average.
In particular, the adaptive method improves the generative performance by up to $19\%$ and $20\%$ compared to the teacher and refining strategy without the adaptive step, respectively.
%\\ \textbf{Effect of oracle accuracy}. 
%The potential bottleneck of our approach is a poor %correlation of existing text-to-image automated estimators %with human preferences.
%For example, ImageReward usually exhibits up to $65 \%$ agreement with annotators. 
%Moreover, it remains unclear what oracle accuracy can be achieved with no-reference decision-making, even if the estimator provides the perfect agreement.
%In \fig{perspectives}, we conduct a synthetic experiment examining the effect of the oracle accuracy on our scheme performance to reveal its future potential. 
% In \fig{perspectives}, we conduct a synthetic experiment examining the effect of the oracle accuracy on our scheme performance to reveal its future potential. 
%if more accurate estimators are developed in future work. 
%We compare the adaptive refinement method ($10$ steps) to SD1.5 ($50$ steps) manually varying the oracle accuracy. 
%We observe significant future gains even for the $75\%$ accuracy rate.
%Note that it remains unclear what level of accuracy can be achieved with no-reference decision-making even if the estimator provides the perfect agreement. %a perfect estimator is available.
\\ \textbf{SDXL results}. 
In addition to SD1.5 experiments, we evaluate our pipeline using the recent CD-SDXL~\cite{luo2023latent} and ADD-XL~\cite{sauer2023adversarial} which are both distilled from the SDXL-Base model~\cite{podell2023sdxl}. 
Our approach with CD-SDXL stands against the top-performing SDXL setting: $50$ steps of the DDIM sampler. 
For ADD-XL, we provide the comparison for $4$ steps where ADD-XL has demonstrated exceptionally strong generative performance in terms of human preference~\cite{sauer2023adversarial}.
In both settings, our approach uses the adaptive refinement strategy with the UniPC solver~\cite{zhao2023unipc}.
Note that both the SDXL-Base and SDXL-Refiner~\cite{podell2023sdxl} models can be used for refinement. 
In our experiments, we observe that the refiner suits slightly better for fixing minor defects while the teacher allows more pronounced changes. 
Thus, we use the refiner for low $\sigma$ values and the base model for the higher ones. %($\leq0.4$)
More setup details are provided in~\ref{app:sdxl_setup}.

The results are presented in \fig{sdxl_res}.
We observe that the adaptive approach using CD-SDXL achieves the quality of the SDXL model, being $5{\times}$ more efficient on average.
Moreover, the proposed scheme improves the performance of ADD-XL by $14\%$ in terms of human preference.

In~\ref{app:diversity}, we also investigate how our approach affects the distribution diversity. \ref{app:oracle_acc} aims to reveal the potential gains of our approach if the oracle accuracy increases in the future.

% \dmitry{todo: For refinement, we use the SDXL Refiner model. Note that the SDXL base model can also be used, but we observe that the refiner performs slightly better for low NFE settings.}
% For both cases, we compare our approach with the top-performing configurations: $50$ steps of the DDIM sampler using the SDXL model and $4$ steps of the ADD-XL. 
% For our approach, we consider the adaptive refinement strategy using UniPC solver~\cite{zhao2023unipc}.
% More details are in~\ref{app:t2i}.
%Note that the ADD-XL overcomes its teacher but suffers from mode collapse~\cite{goodfellow2020generative}. However, this is not the case for the CD~\cite{song2023consistency}, as we discuss in Section~\ref{sec:diversity}.
%we perform $4$ sampling steps with the student model and consider $12$ steps for the refinement strategy.
% As a result, the adaptive scheme with $70\%$ acceptance rate consumes $13$ NFE in total and we compare it with 
%Also, we compare our approach against ADD-XL using $4$ generation steps which has demonstrated higher human preferences than the teacher model~\cite{sauer2023adversarial}.
%In the adaptive setting, we consider $2$ steps of ADD-XL, $4$ refining steps and $50\%$ acceptance rate, resulting in $4$ NFE.
%In both evaluations, the guidance scale is $8$ for all models and SDXL Refiner~\cite{podell2023sdxl} is used for the refining strategy.

\begin{table}[t!]
\centering
\resizebox{0.95\linewidth}{!}{%
\begin{tabular}{c c c c c } 
\bottomrule
{Method} & Steps & DINOv2 $\downarrow$ & ImageReward $\uparrow$\\
\hline 
\iffalse
CD  &$5$ & $0.674$ \small{$\pm \ .004$} &  $0.192$ \small{$\pm \ .037$} \\ 
DDIM  &$50$& $0.665$ \small{$\pm \ .007$}  &  $0.183$ \small{$\pm \ .024$} \\ 
DDIM &$25$ & $0.665$ \small{$\pm \ .002$} &  $0.183$ \small{$\pm \ .022$} \\ 
DPM &$25$ & $0.667$ \small{$\pm \ .005$} &  $0.179$ \small{$\pm \ .020$} \\ 
Refinement &$30$ & $0.710$ \small{$\pm \ .005$} & $0.383$ \small{$\pm \ .033$}  \\ 
\hline
Ours &$30$ & ${0.669}$ \small{$\pm \ .006$} &  ${0.281}$ \small{$\pm \ .008$}  \\
% Ours (IR + DINOv2) & $\mathbf{0.336}$ \small{$\pm \ .005$} &  $\mathbf{0.255}$ \small{$\pm \ .022$}  \\
\fi
CD-SD1.5         & 5  & \cellcolor[HTML]{f0ffcc}0.674 \small{$\pm \ $ .004} & \cellcolor[HTML]{ffcccd}0.192 \small{$\pm \ $.037} \\ 
SD1.5, DDIM       & 50 & \cellcolor[HTML]{ccffcc}0.665 \small{$\pm \ $ .007} & \cellcolor[HTML]{ffcccd}0.183 \small{$\pm \ $.024} \\ 
SD1.5, DDIM       & 25 & \cellcolor[HTML]{ccffcc}0.665 \small{$\pm \ $ .002} & \cellcolor[HTML]{ffcccd}0.183 \small{$\pm \ $.022} \\ 
SD1.5, DPM        & 25 & \cellcolor[HTML]{ccffcc}0.667 \small{$\pm \ $ .005} & \cellcolor[HTML]{ffcccd}0.179 \small{$\pm \ $.020} \\ 
Refinement & 30 & \cellcolor[HTML]{ffcccd}0.710 \small{$\pm \ $.005}  & \cellcolor[HTML]{ccffcc}0.383 \small{$\pm \ $.033}  \\ 
\hline
Ours     & 30 & \cellcolor[HTML]{ccffcc}0.669 \small{$\pm \ $.006} &  \cellcolor[HTML]{ffffcc}0.281 \small{$\pm \ $.008}  \\
\bottomrule
\end{tabular} 
}
\caption{Comparison of SDEdit using different approaches in terms of reference preservation and editing quality for the strength $0.6$.}
\label{tab:editing_fixed}
\vspace{-3mm}
\end{table}

\subsection{Text-guided image editing}
This section applies our approach for text-guided image editing using SDEdit~\cite{meng2021sdedit}. 
We add noise to an image, alter the text prompt and denoise it using the student model first.
If the edited image does not exceed the threshold $\tau$, the teacher model is used for editing instead.
In the editing setting, we observe that the refinement strategy significantly reduces similarity with the reference image due to the additional noising step. 
Thus, we apply the regeneration strategy only.

\begin{figure}[t!]
    \centering
    \includegraphics[width=1.0\linewidth]{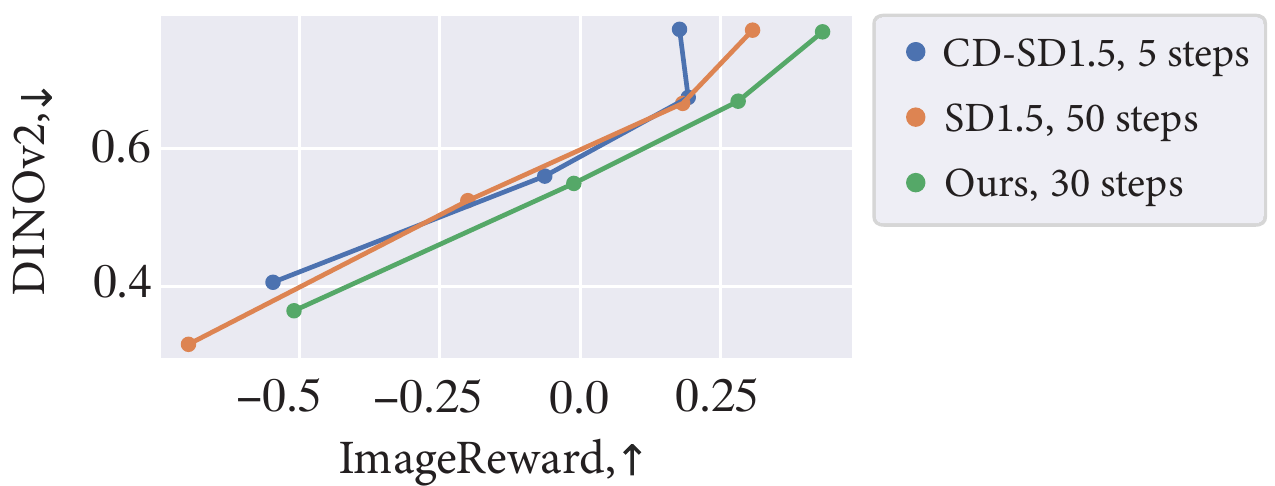}
    \vspace{-6mm}
    \caption{SDEdit performance for different strength values in terms of reference preservation (DINOv2) and editing quality (IR).} %trade-off curves for different strength values in $[0.3, 0.9]$.}
    \label{fig:editing_curve}
\end{figure}

In these experiments, the SD1.5 and CD-SD1.5 models are considered.
As performance measures, we use ImageReward for editing quality and DINOv2~\cite{oquab2023dinov2} for reference preservation. 
For evaluation, $100$ text prompts from COCO2014 are manually prepared for the editing task. 
% Other settings are similar to Section \ref{subsec:image_synth}. 
\\ \textbf{Results}. 
\tab{editing_fixed} provides evaluation results for a SDEdit noising strength value $0.6$.
The proposed method demonstrates a higher ImageReward score compared to the baselines with similar reference preservation scores. 
% We observe that the refinement strategy significantly reduces similarity with the original image due to the additional noising step. 
% Thus, in these experiments, we consider the adaptive regeneration strategy only. 
% Secondly, we evaluate different estimators, including ImageReward and a combination of ImageReward and DINOv2. 
% We find that the second approach preserve the reference slightly better. 
In addition, we present the performance for different editing strength values in Figure \ref{fig:editing_curve}. 
Our approach demonstrates a better trade-off between reference preservation and editing quality. We provide qualitative results in Figure \ref{fig:ours_teacher}.

\subsection{Controllable image generation}
Finally, we consider text-to-image generation using Canny edges and semantic segmentation masks as an additional context and use ControlNet~\cite{zhang2023adding} for this task.
We use ControlNet pretrained on top of SD1.5 and directly plug it into the distilled model (CD-SD1.5).
Interestingly, the model pretrained for the teacher model fits the student model surprisingly well without any further adaptation.

For the teacher model, the default ControlNet sampling configuration is used: $20$ sampling steps of the UniPC~\cite{zhao2023unipc} solver. 
In our adaptive approach, we use the refinement strategy with $10$ steps of the same solver.
For performance evaluation, we conduct the human preference study for each task on $600$ examples and provide more details in \ref{app:applications}.
\\ \textbf{Results.} 
According to the human evaluation, our approach outperforms the teacher ($20$ steps) by $19 \%$ ($9$ steps) and $4 \%$ ($11$ steps) for Canny edges and semantic segmentation masks, respectively.
The visual examples are in Figure~\ref{fig:ours_teacher}.

%\begin{table}[t!]
%\centering
%\resizebox{\linewidth}{!}{
%\begin{tabular}{c c | c c c | c} 
%\toprule
%Method & NFE & Viewpoint & Background & Main object  %& Avg\\
%\midrule
%CD-SDXL & $4$ & $0.547$ & $0.627$ & $0.753$ & %\cellcolor[HTML]{ffffcc}$0.642$\\
%Adaptive (R) & $9$ & $0.581$ & $0.666$ & $0.757$  & %\cellcolor[HTML]{f0ffcc}$0.668$\\
%Adaptive (G) & $29$ & $0.629$ & $0.604$ & $0.766$ & %$0.666$\\
%SDXL & $50$ & $0.670$ & $0.593$ & $0.762$ & %\cellcolor[HTML]{ccffcc}$0.675$ \\
%\midrule
%ADD-XL & $4$ & $0.314$ & $0.531$ & $0.637$ & %\cellcolor[HTML]{ffcccd}$0.494$\\
%Adaptive (R) & $4$ & $0.434$ & $0.476$ & $0.672$  & %$0.527$\\
%Adaptive (G) & $29$ &$0.471$ & $0.554$ & $0.665$ & %$0.563$\\
%SDXL & $50$ &$0.670$ & $0.593$ & $0.762$  & %\cellcolor[HTML]{ccffcc}$0.675$ \\
%\midrule
%CD-SD1.5 & $5$ &$0.620$ & $0.678$ & $0.918$ & %\cellcolor[HTML]{ccffcc}$0.738$\\
%Adaptive (R) & $12$ & TODO & TODO & TODO & TODO\\
%Adaptive (G) & $40$ & TODO & TODO & TODO  & TODO\\
%SD1.5 & $50$ & $0.658$ & $0.667$ & $0.917$ & %\cellcolor[HTML]{ccffcc}$0.747$\\
%\bottomrule
%\end{tabular}
%}
%\caption{.}
%\label{table:diversity}
%\end{table}

%% file: sec/6_conclusion.tex
\section{Conclusion}
This work investigates the performance of the distilled text-to-image models and demonstrates that they may consistently outperform the teachers on many samples.
We design an adaptive text-to-image generation pipeline that takes advantage of successful student samples and, in combination with the teacher model, outperforms other alternatives for low and high inference budgets.  
%Importantly, text-to-image automated estimators and distillation methods still have room for further development.
%Therefore, we believe the proposed approach may be even in higher demand when future works make a significant step forward in these directions.

%\dmitry{TODO: discuss that there are no other publicly available very cheap text-to-image proxies instead of distilled models. It owuld be insteresting to consider other collaborations of the very cheap proxy and expensive models (not teacher and student) in future work, e.g., T2I gans and diffusions.}

%% file: sec/X_suppl_camera.tex
\clearpage
\onecolumn

\setcounter{page}{1}

\newpage
\appendix
\maketitlesupplementary

\section{CD-SD1.5 implementation details}
\label{app:student_details}
In this work, we develop consistency distillation (CD) for Stable Diffusion following the official implementation~\cite{song2023consistency}. 
For training, we prepare a subset of LAION2B~\cite{schuhmann2022laion}, which consists of $80$M image-text pairs. 
As a teacher sampler, we consider DDIM-solver using $50$ sampling steps and Variance-Preserving scheme~\cite{song2020score}. 
We use the teacher UNet architecture as a student model and initialize it with the teacher parameters. 
Classifier-free guidance is applied to the distilled model directly without merging it into the model as done in~\cite{meng2023distillation}. %the use of an additional distillation step
%We leave it for the future work. 
During training, we uniformly sample the guidance strength from $1$ to $8$. 
Thus, our model supports different guidance scales during sampling. 
We train the student for ${\sim}200$K iterations on $8$ A100 GPUs using the following setting: $512$ batch size; $0.95$ EMA rate; $1\text{e}{-}5$ fixed learning rate; L2 uniformly weighted distillation loss calculated in the latent space of the VAE encoder. 
During inference, the multistep stochastic sampler~\cite{song2023consistency} is used to generate images. 
In most of our experiments, we use $5$ sampling steps.

Note that we use our implementation of consistency distillation for SD because, when most experiments were conducted, there were no publicly available implementations.

\begin{figure*}[t!]
    \centering
    \vspace{-2mm}
    \includegraphics[width=0.95\linewidth]{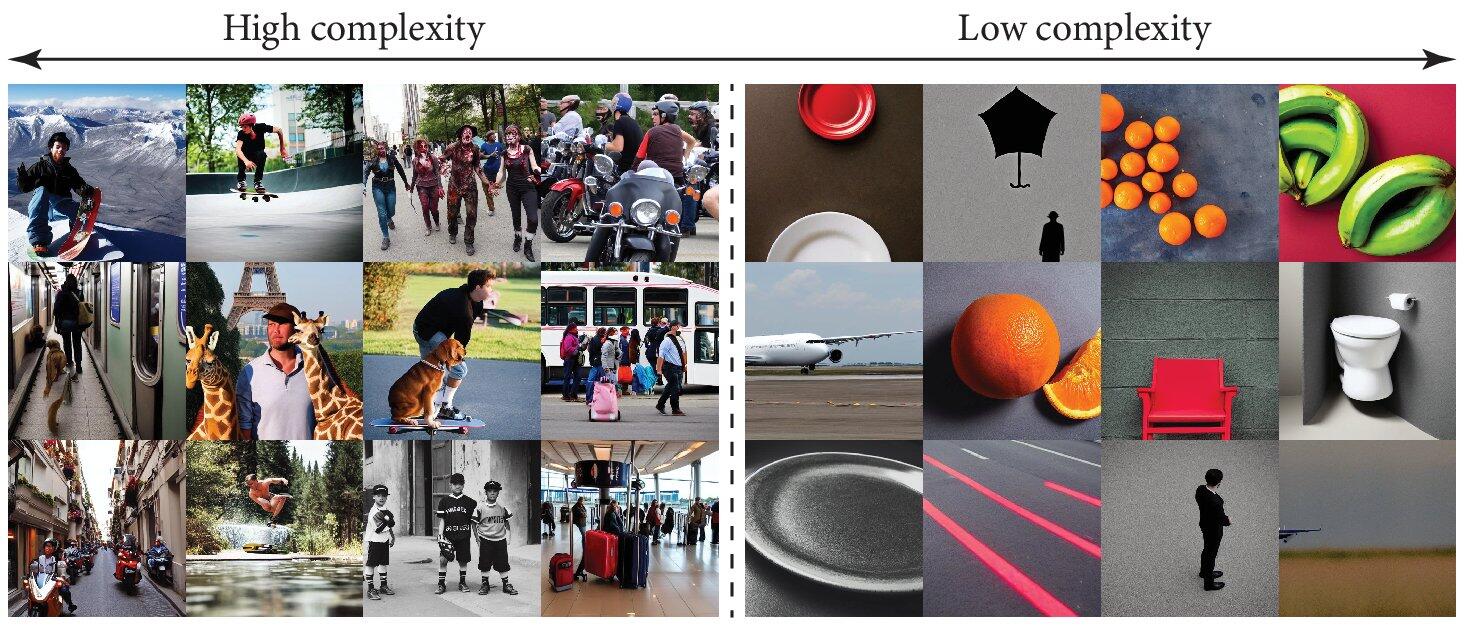}
    \vspace{-3mm}
    \caption{SD1.5 samples of different complexity according to the ICNet model~\cite{feng2023ic9600}.}
    \label{fig:app_complexity}
\end{figure*}

\begin{figure*}[t!]
    \centering
    \includegraphics[width=\linewidth]{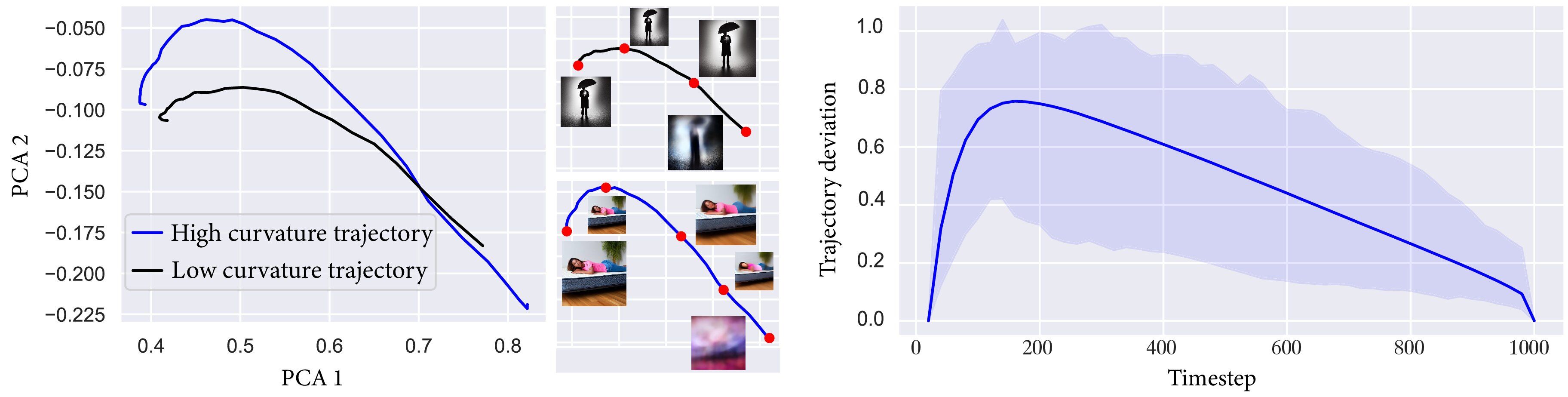}
    \vspace{-6mm}
    \caption{\textit{Left}: Two examples of diffusion model trajectories with high and low curvatures. \textit{Right}: Trajectory deviations according to~\cite{chen2023geometric}.}
    \label{fig:app_curv}
\end{figure*}

\section{Analysis} \label{app:analysis}
\subsection{Details}
\label{app:calc_details}
\textbf{Image complexity}. 
To calculate the image complexity, we use the recent ICNet model~\cite{feng2023ic9600}. 
This model is learned on a large-scale human annotated dataset. 
Each image corresponds to a complexity score ranging from 0 (the simplest) to 1 (the most complex). 
In Figure~\ref{fig:app_complexity}, we provide examples of Stable Diffusion samples with the lowest and highest image complexity. 
More complex images usually depict multiple entities, often including people, and intricate backgrounds.
\\ \textbf{Text influence}. 
We calculate the influence of a text prompt on student generation by using cross-attention between token embeddings and intermediate image representations. 
Following~\cite{tang2023daam}, we collect cross-attention maps for all diffusion steps and UNet~\cite{ronneberger2015u} layers. 
Then, the average attention score is calculated for each text token. 
Finally, the highest value among all tokens is returned.
\\ \textbf{Trajectory curvature} is estimated according to the recent work~\cite{chen2023geometric}. 
First, we calculate the trajectory deviations as L2 distance from the denoised prediction at a time step $t$ to the straight line passing
through the denoising trajectory endpoints. 
The trajectory curvature corresponds to the highest deviation over all time steps. 

In Figure~\ref{fig:app_curv} (Left), we visualize two denoising trajectories corresponding to high and low curvatures. 
We apply PCA~\cite{jolliffe2016principal} to reduce the dimensionality of the denoised predictions.
In addition, Figure~\ref{fig:app_curv} (Right) demonstrates trajectory deviations for different time steps. 
The highest deviations typically occur closer to the end of the denoising trajectory.

%\begin{figure*}
%    \centering
%    \includegraphics[width=\linewidth]{images/app_student_better.jpg}
%    \caption{Student outperforms its teacher according to the human evaluation.}
%    \label{fig:app_student_better2}
%\end{figure*}

\subsection{Other distilled text-to-image models}
\label{app:other_distill}
Here, we conduct a similar analysis as in Section \ref{main_analysis} for consistency distillation on Dreamshaper v$7$~\cite{luo2023latent} and architecture-based distillation\footnote{\url{https://huggingface.co/docs/diffusers/using-diffusers/distilled_sd}}.

In contrast to the few-step distillation approaches~\cite{meng2023distillation, song2023consistency, kim2023consistency, liu2022flow, luo2023latent}, the architecture-based distillation removes some UNet layers from the student model for more efficient inference and trains it to imitate the teacher using the same deterministic solver and number of sampling steps. 

\begin{figure*}
    \centering
    \begin{subfigure}[b]{\linewidth}
         \centering
         \includegraphics[width=0.88\linewidth]{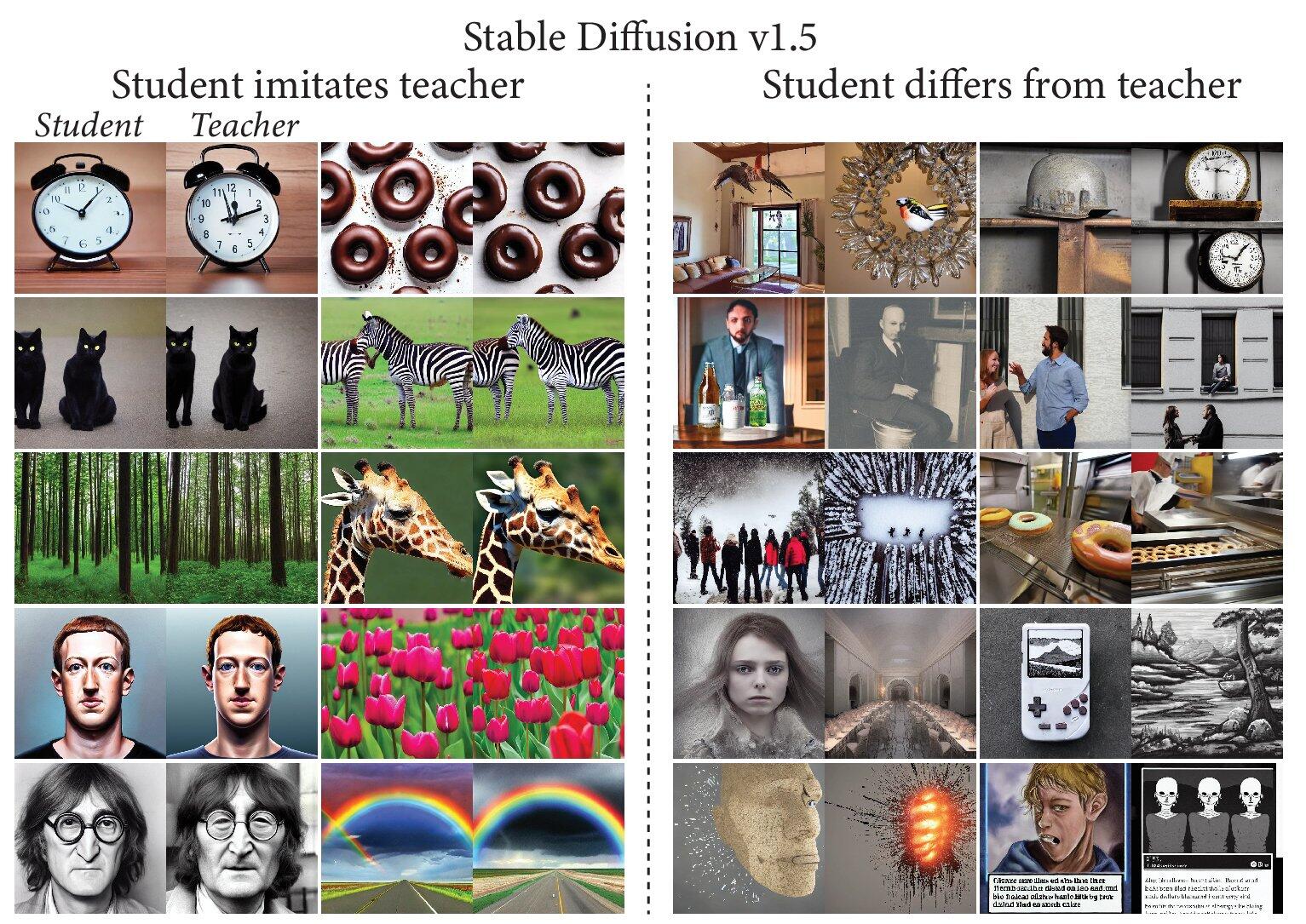}
         \caption{}
         \label{fig:app_dual_mode_stable}
     \end{subfigure}
    \begin{subfigure}[b]{\linewidth}
         \centering
         \includegraphics[width=0.88\linewidth]{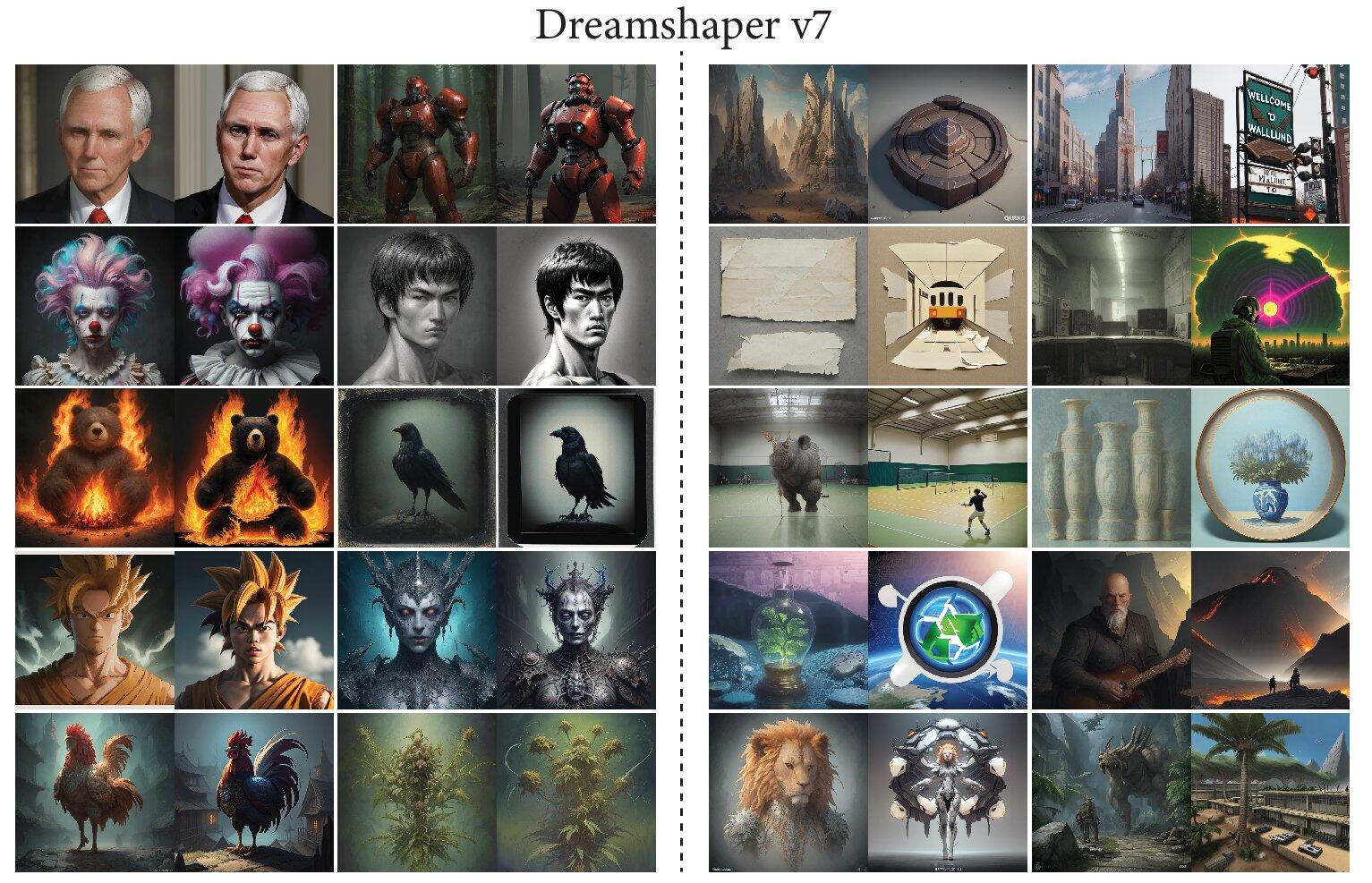}
         \caption{}
         \label{fig:app_dual_mode_dream}
     \end{subfigure}
     \caption{Visual examples of similar (Left) and dissimilar (Right) teacher and student samples for SD1.5 (a) and Dreamshaper v7 (b).}
     \label{fig:app_dual_mode_dream_stable}
\end{figure*}

\begin{figure*}[t!]
    \centering
    \includegraphics[width=1.05\linewidth]{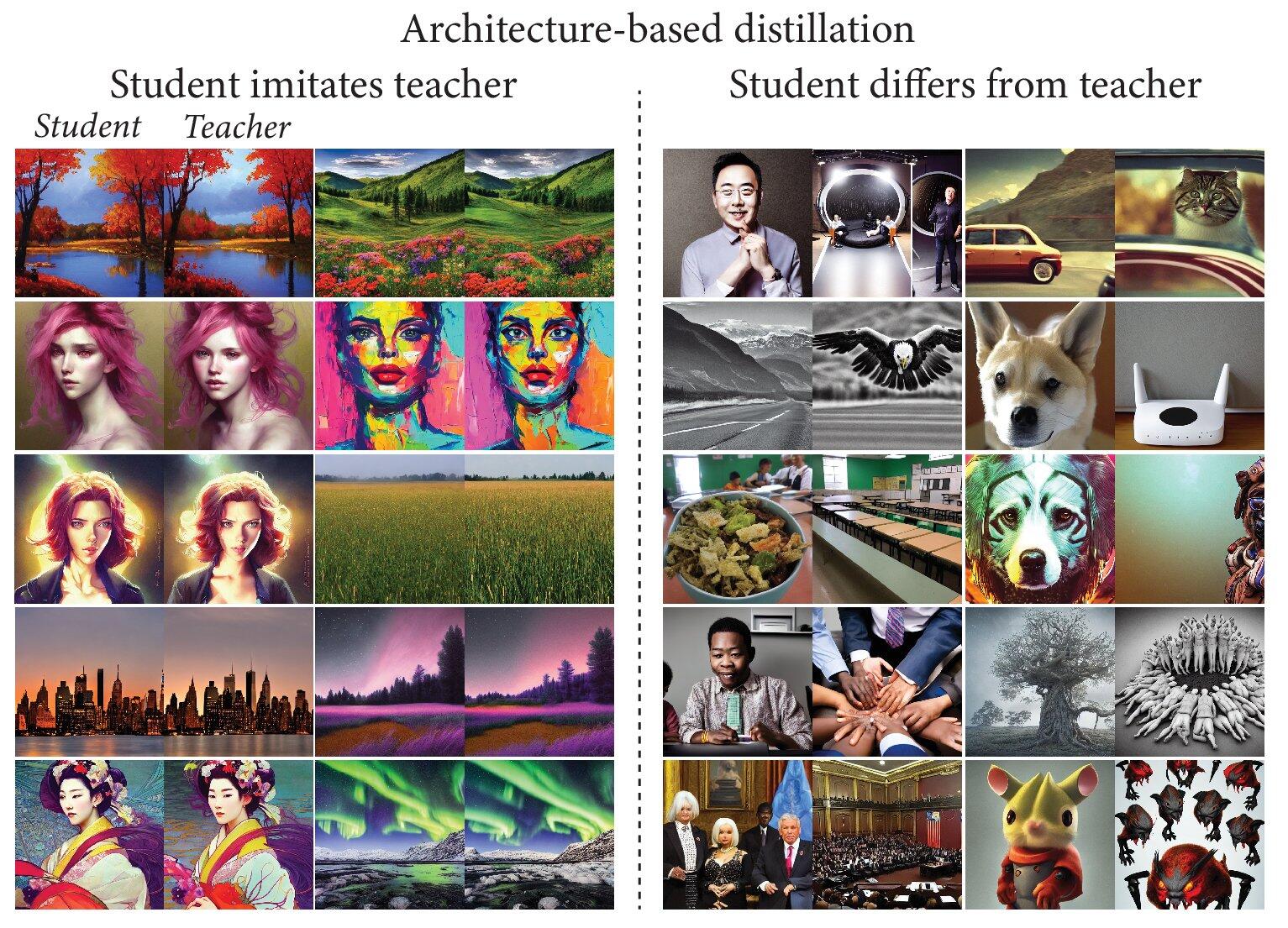}
    \caption{Visual examples of similar (Left) and dissimilar (Right) teacher and student samples for the architecture-based distillation.}
    \label{fig:app_dual_mode_arch}
\end{figure*}

\begin{figure*}[t!]
    \centering
    \includegraphics[width=\linewidth]{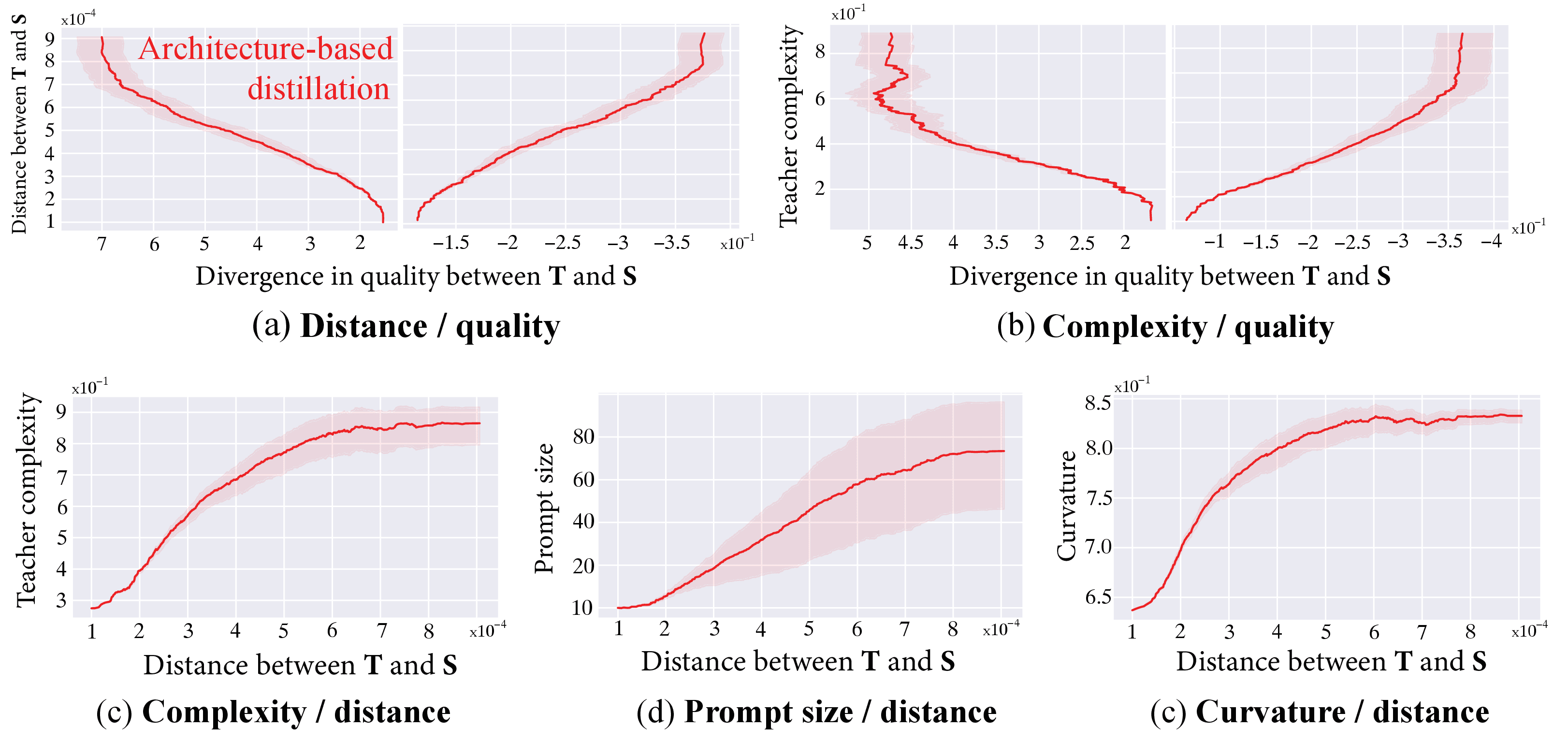}
    \caption{Analysis results for the architecture-based distillation.}
    \label{fig:app_all_arch}
\end{figure*}

\begin{figure*}[t!]
    \centering
    \includegraphics[width=\linewidth]{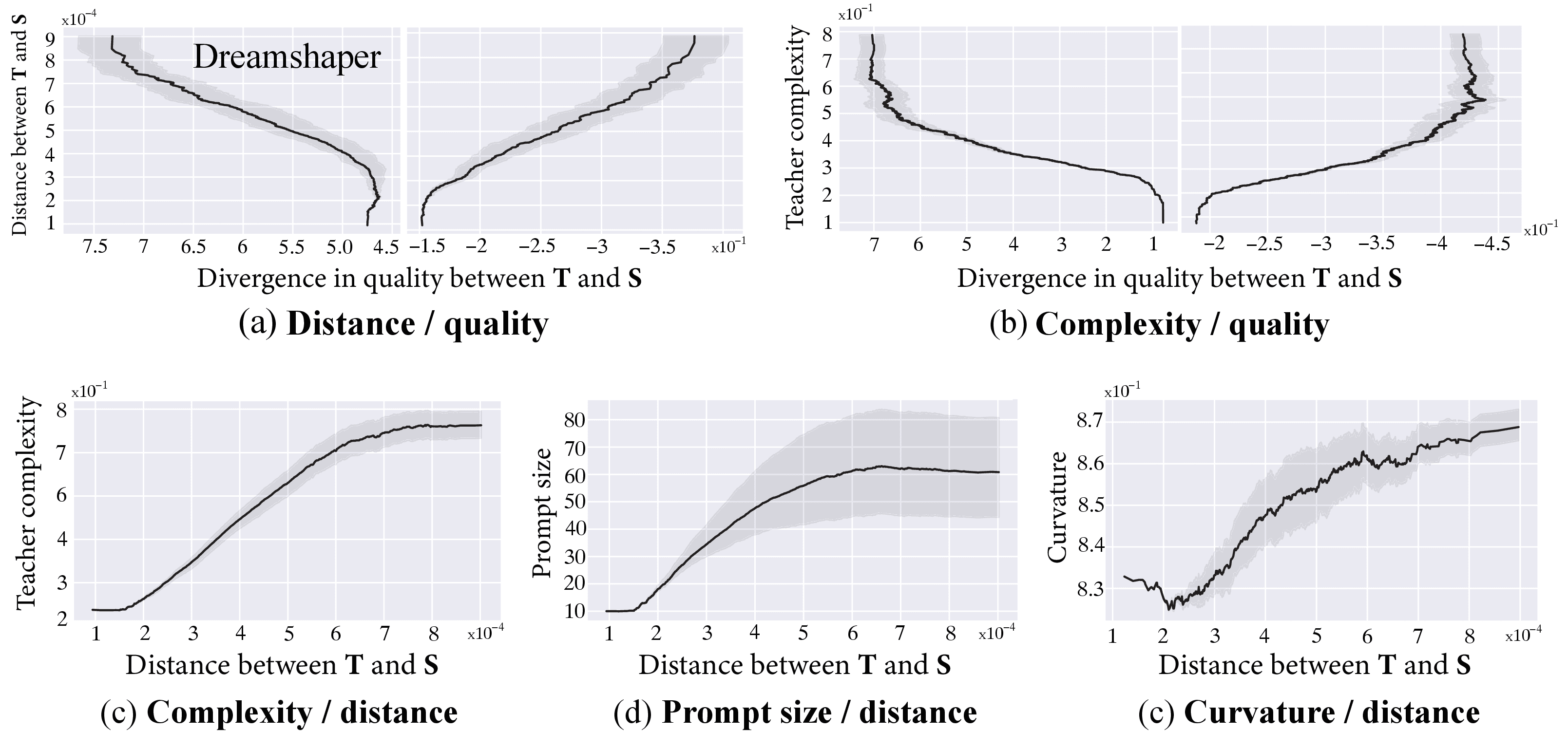}
    \caption{Analysis results for the consistency distillation on Dreamshaper v7.}
    \label{fig:app_all_lcm}
\end{figure*}

In Figures~\ref{fig:app_dual_mode_dream_stable},~\ref{fig:app_dual_mode_arch}, we show that both methods can produce samples that significantly differ from the teacher ones for the same text prompt and initial noise. 
Then, Figures~\ref{fig:app_all_arch},~\ref{fig:app_all_lcm} confirm that other observations in Section~\ref{main_analysis} remain valid for the Dreamshaper and architecture-based students as well.

% \section{Method} 
% \label{app:method}
\begin{algorithm}[t!]
\DontPrintSemicolon
  \KwInput{$\mathbf{E}, \mathbf{S}, \mathbf{T} - $ estimator, student, teacher; \\ $\mathcal{I} - $ input; $\sigma, \tau - $ rollback value and cut-off threshold.}
  \vspace{1.5mm}
  $\hat{\mathcal{O}} = \mathbf{S}(\mathcal{I})$ \tcp*{Student prediction}
  \vspace{1.5mm}
  \If{$\mathbf{E}(\hat{\mathcal{O}}) < \tau$ } 
  {
    \vspace{1.5mm}
    \textit{Strategy 1}: \tcp*{Refinement}
       \ \ \ \ $\hat{\mathcal{O}}_{\sigma} = \sqrt{1 - \sigma} \cdot \hat{\mathcal{O}} + \sqrt{\sigma} \cdot \mathcal{Z}, \ \mathcal{Z} \sim \mathcal
       {N}(0,1)$ \\
       \ \ \ \ $\hat{\mathcal{O}} = \mathbf{T}(\mathcal{I}, \hat{\mathcal{O}}_{\sigma})$ \\
    \textit{Strategy 2}: \tcp*{Regeneration}
        \ \ \ \ $\hat{\mathcal{O}} = \mathbf{T}(\mathcal{I})$ 
  }
  {\Return{$\hat{\mathcal{O}}$}}
\caption{teacher-student adaptive collaboration}
\label{alg:method_alg}
\end{algorithm}

% \newpage
\section{Cut-off threshold tuning}
\label{app:method}
%\subsection{Rollback value}
%Our approach uses two primary hyperparameters: $\tau$ (cut-off threshold) and $\sigma$ (rollback value). 
%The rollback value controls the corruption strength in the adaptive refinement stage and can be varied between $0$ and $1$. 
%$1$ corresponds to the complete noise without any %information about an original image and $0$ to an initial %image without noise.
%Both hyperparameters are tuned on a hold-out subset for a %specified metric, e.g., ImageReward. 

The cut-off threshold $\tau$ is used for the adaptive selection of student samples. 
It corresponds to the $k$-th percentile of the metric values calculated on validation student samples. 
$600$ and $300$ samples are generated for tuning on the COCO2014 and LAION-Aesthetics datasets, respectively. 
Note that the prompts used for tuning do not overlap with the test ones.
Then, we calculate an individual score, e.g., IR score, for each validation student sample and select the percentile based on an average inference budget or target metric value. 
For example, suppose we select the percentile given a $15$ step budget and intend to perform $5$ student steps and $20$ steps for the improvement strategy.
In this case, we have to select $\tau$ as a $50$-th percentile, which results in the final average number of steps: $5+0.5\cdot 20=15$.

During inference, we perform the adaptive selection as follows: if the score of the student sample exceeds $\tau$, 
we consider that this sample might be superior to the teacher one and keep it untouched. 
Otherwise, we perform an improvement step using the teacher model (refinement or regeneration).
The proposed pipeline is presented in Algorithm~\ref{alg:method_alg}.

We also show that hyperparameter tuning is straightforward and requires negligible effort. 
To verify this, we tune the threshold using the various number of prompts (Tab.~\ref{tab_1}). 
 We can see that $500$ prompts are sufficient for the threshold convergence.
 %We can see that $500$ prompts are sufficient for tuning, after which the threshold reaches a plateau. 
 Thus, the process only needs to generate $500$ student samples and takes ${\sim}2.5$ minutes.
 \begin{table}[h!]
\centering
\resizebox{\linewidth}{!}{%
\begin{tabular}{@{} ll *{6}{c} @{}}
\toprule
Prompts  & $150$ & $250$  & $500$ & $5000$ & $25000$\\
\midrule
    Diffusion-DB & $0.646 \  \scriptstyle{.104}$ & $0.565 \  \scriptstyle{.028}$  & $0.569 \  \scriptstyle{.010}$ & $0.570 \  \scriptstyle{.008}$ & $0.568 \  \scriptstyle{.004}$  \\
 PickScore &  $0.537 \  \scriptstyle{.032}$  & $0.436 \  \scriptstyle{.019}$ & $0.479 \  \scriptstyle{.008}$ & $0.476 \  \scriptstyle{.006}$  & $0.481 \  \scriptstyle{.002}$  \\
\midrule
 Time, sec. & $44.9 \  \scriptstyle{.4}$  & $73.8 \  \scriptstyle{.2}$ & $148 \  \scriptstyle{2}$ & $1470 \  \scriptstyle{8}$  & $7448 \  \scriptstyle{10}$  \\
\bottomrule
\end{tabular} }
\caption{The threshold value tuned using prompts from Diffusion-DB and PickScore datasets and the time required for tuning.}
\label{tab_1}
\end{table}

\section{Experiments} 
\subsection{Human evaluation}
\label{app:human_eval}
To evaluate the text-to-image performance, we use the side-by-side comparison conducted by professional annotators. 
Before the evaluation, all annotators pass the training and undergo the preliminary testing. 
Their decisions are based on the three factors: textual alignment, image quality and aesthetics (listed in the priority order). 
Each side-by-side comparison is repeated three times by different annotators. 
The final result corresponds to the majority vote.

\subsection{Experimental setup (SD1.5)}
\label{app:sd15_setup}
% In this case, we use from $3$ to $5$ steps of the CD-SD1.5. 
% Then, we use a teacher model (SD 1.5) with DPM-Solver~\cite{luu2022dpm}, adaptively applying from $5$ to $45$ steps.
The exact hyperparameter values and number of steps used for the automated estimation (FID, CLIP score and ImageReward) of the adaptive refinement strategy in Table~\ref{app:tab_hypers}. 
The adaptive regeneration uses the same cut-off thresholds and number of steps, but the rollback value, $\sigma$, is equal to $1$.
The values used for human evaluation are presented in Table~\ref{app:tab_hypers_human}.

\subsection{Experimental setup (SDXL)}
\label{app:sdxl_setup}
We evaluate two distilled models: CD-SDXL~\cite{luo2023latent} and ADD-XL~\cite{sauer2023adversarial}. 
For the first evaluation, we use $4$ steps of the CD-SDXL and then apply $12$ adaptive refinement steps using the teacher model (SDXL-Base~\cite{podell2023sdxl}) with the UniPC solver~\cite{zhao2023unipc}. 
We compare our pipeline to the default teacher configuration: $50$ DDIM steps. 
For the second evaluation, we perform $2$ steps of the ADD-XL and $4$ steps of the SDXL-Refiner~\cite{podell2023sdxl} for the adaptive refinement strategy.
We compare to $4$ ADD-XL steps as this setting outperformed SDXL-Base in terms of image quality and textual alignment~\cite{podell2023sdxl}.
The exact hyperparameter values and number of steps used for human evaluation are in Table~\ref{app:tab_hypers_human_sdxl}.

We found that SDXL-Refiner performs slightly better than the base model for small refinement budgets (e.g., $4$). 
The refiner typically helps to improve fine-grained details, e.g., face attributes or background details.
However, it faces difficulties in providing global changes and sometimes brings artifacts for large rollback values, $\sigma$. 
Thus, we use the SDXL-Base teacher for more refinement steps (e.g., 12).

\begin{table}[h!]
\centering
\resizebox{\linewidth}{!}{%
\begin{tabular}{@{} ll *{6}{c} @{}}
\toprule
Metric & $\sigma$
  & $k$
  & \multicolumn{3}{c}{Steps}\\
\cmidrule(l){4-6}
 & & &CD & Refinement & Adaptive\\
\midrule
ImageReward & $0.4$  & $60$ & $5$ & $5$  & $8$ \\
 & $0.55$  & $60$ & $5$ & $10$  & $11$ \\
 & $0.7$  & $60$ & $5$ & $15$  & $14$ \\
 & $0.7$  & $60$ & $5$ & $25$  & $20$ \\
 & $0.7$  & $60$ & $5$ & $35$  & $26$ \\
 & $0.7$  & $60$ & $5$ & $45$  & $32$ \\
\midrule
CLIP score & $0.5$  & $60$ & $5$ & $5$  & $8$ \\
 & $0.7$  & $60$ & $5$ & $10$  & $11$ \\
 & $0.75$  & $60$ & $5$ & $15$  & $14$ \\
 & $0.75$  & $60$ & $5$ & $25$  & $20$ \\
 & $0.75$  & $60$ & $5$ & $35$  & $26$ \\
 & $0.75$  & $60$ & $5$ & $45$  & $32$ \\
\midrule
FID & $0.75$  & $40$ & $3$ & $5$  & $5$ \\
 & $0.7$  & $70$ & $3$ & $15$  & $14$ \\
\bottomrule
\end{tabular} }
\vspace{-2mm}
\caption{Hyperparameter values used for the \textbf{automated evaluation (SD 1.5)}, Figure \ref{fig:automatic_t2i}.}
\label{app:tab_hypers}
\end{table}

\begin{table}[h!]
\centering
\resizebox{\linewidth}{!}{%
\begin{tabular}{@{} ll *{6}{c} @{}}
\toprule
Metric & $\sigma$
  & $k$
  & \multicolumn{3}{c}{Steps}\\
\cmidrule(l){4-6}
 & & &CD &Refinement &Adaptive\\
\midrule
Human evaluation & $0.7$  & $50$ & $5$ & $10$  & $10$ \\
 & $0.7$  & $50$ & $5$ & $20$  & $15$ \\
 & $0.7$  & $50$ & $5$ & $40$  & $25$ \\
\bottomrule
\end{tabular} }
\vspace{-2mm}
\caption{Hyperparameter values used for the \textbf{user preference study (SD 1.5)}, Figure \ref{fig:users_study}.}
\label{app:tab_hypers_human}
\end{table}

\begin{table}[h!]
\centering
\resizebox{\linewidth}{!}{%
\begin{tabular}{@{} ll *{6}{c} @{}}
\toprule
Metric & $\sigma$
  & $k$
  & \multicolumn{3}{c}{Steps}\\
\cmidrule(l){4-6}
 & & & ADD-XL &Refinement &Adaptive\\
\midrule
Human evaluation & $0.4$  & $50$ & $2$ & $4$  & $4$ \\
\toprule \toprule
&  & & CD-SDXL &Refinement &Adaptive\\
\midrule
Human evaluation & $0.85$  & $70$ & $4$ & $12$  & $13$ \\
\toprule
\end{tabular}  } 
\vspace{-2mm}
\caption{Hyperparameter values used for the \textbf{user preference study (SDXL)}, Figure \ref{fig:sdxl_res}.}
\label{app:tab_hypers_human_sdxl}
\end{table}

\subsection{Effect of oracle accuracy}
\label{app:oracle_acc}
\begin{figure}[t!]
    \centering
    \includegraphics[width=1.0\linewidth]{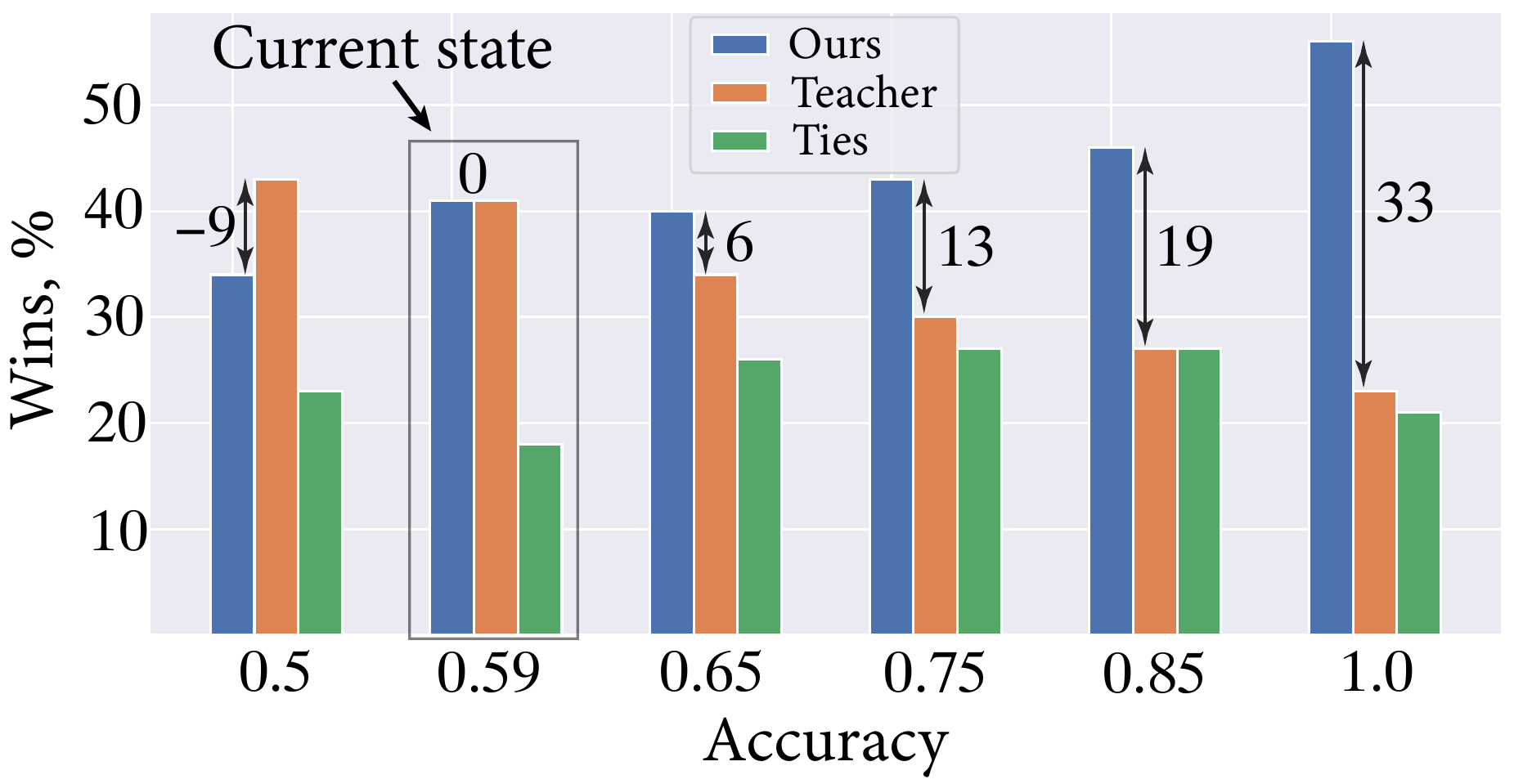}
    \vspace{-6mm}
    \caption{User preferences for different accuracy levels of the no-reference decision-making procedure.
    the automated sample estimator. 
    \textit{Current state} represents the results using ImageReward. 
    The results for higher accuracy rates demonstrate the future gains if the oracle performance improves.}
    \label{fig:perspectives}
\end{figure} 
The potential bottleneck of our approach is a poor correlation of existing text-to-image automated estimators with human preferences.
For example, ImageReward usually exhibits up to $65 \%$ agreement with annotators. 
Moreover, it remains unclear what oracle accuracy can be achieved with no-reference decision-making, even if the estimator provides the perfect agreement.
In \fig{perspectives}, we conduct a synthetic experiment examining the effect of the oracle accuracy on our scheme performance to reveal its future potential. 
% if more accurate estimators are developed in future work. 
We compare the adaptive refinement method ($10$ steps) to SD1.5 ($50$ steps) manually varying the oracle accuracy. 
We observe significant future gains even for the $75\%$ accuracy rate.
%Note that it remains unclear what level of accuracy can be achieved with no-reference decision-making even if the estimator provides the perfect agreement. %a perfect estimator is available.

\subsection{Distribution diversity} \label{app:diversity}
\begin{figure}[t!]
    \centering
    \vspace{-1mm}
    \includegraphics[width=0.99\linewidth]{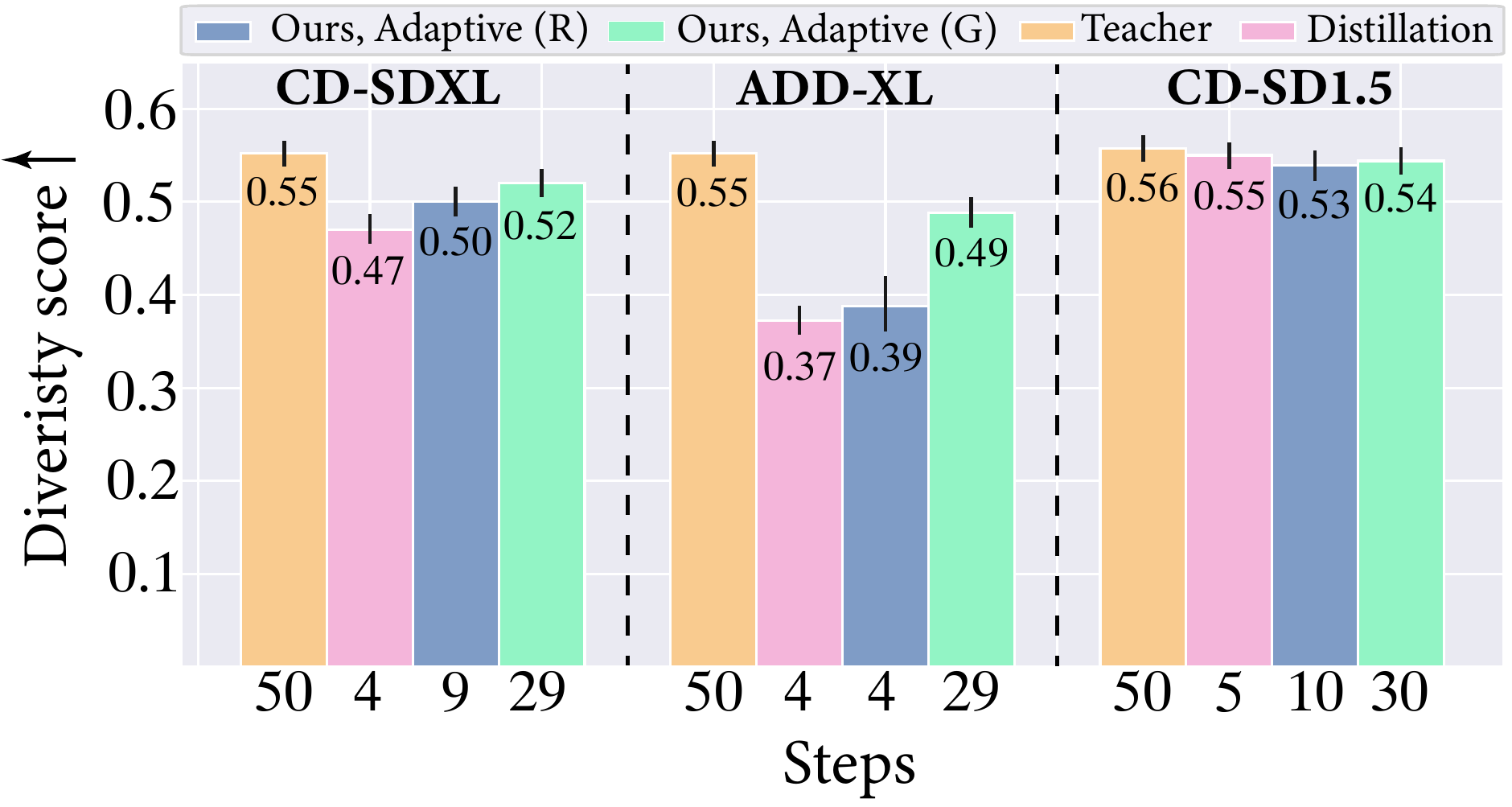}
    \vspace{-3mm}
    \caption{Diversity human scores collected for different methods.}
    \label{fig:diversity_scores}
\end{figure}
\begin{figure}[t!]
    \centering
    \includegraphics[width=\linewidth]{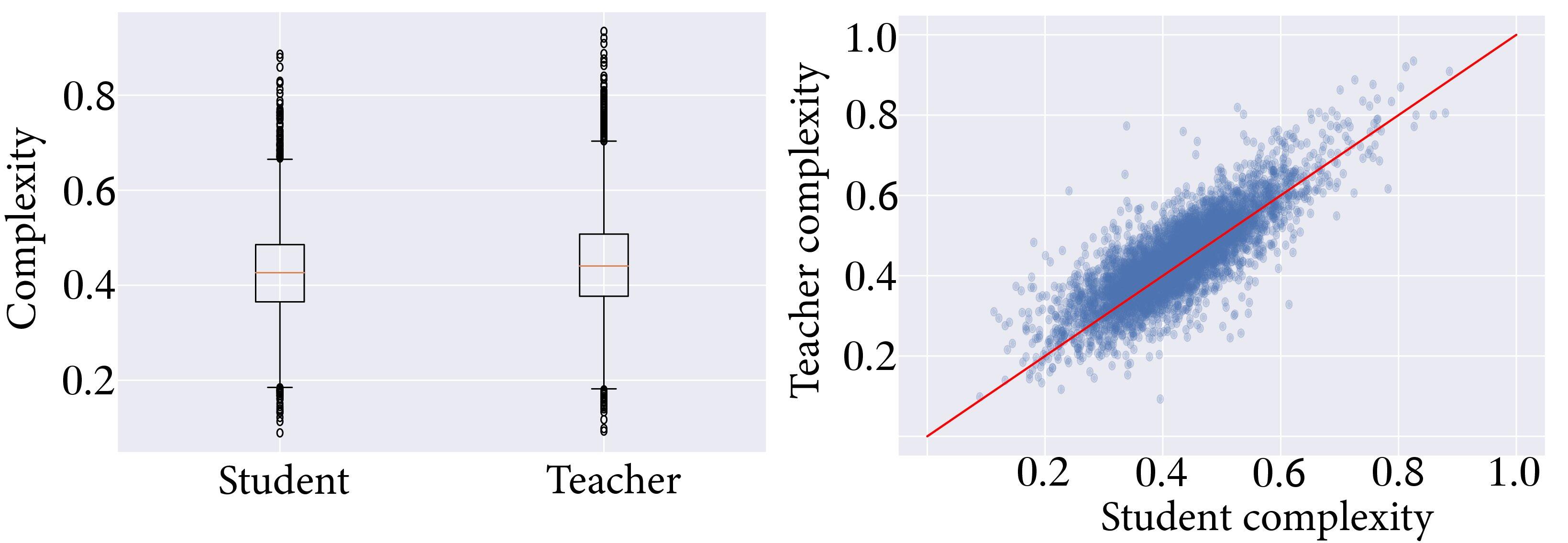}
    \vspace{-6mm}
    \caption{Image complexity of the CD-SD1.5 and SD1.5 samples in terms of ICNet~\cite{feng2023ic9600}. \textit{Left:} Box plot representing the complexity quantiles for both models. \textit{Right:} Distribution of individual complexity values. Each dot corresponds to a pair of samples generated for the same prompt and initial noise. The distilled model only slightly simplifies the teacher distribution.}
    \label{fig:app_student_compl}
    \vspace{-1mm}
\end{figure}

In the proposed teacher-student collaboration, the oracle aims to accept high-quality and well-aligned student samples but does not control the diversity of the resulting image distribution.
Therefore, if the student exhibits severe mode collapse or oversimplifies the teacher samples, the adaptive pipeline will likely inherit these issues to some extent.

In this section, we investigate this potential problem for several existing distilled text-to-image models. 
Specifically, we consider consistency distillation~\cite{luo2023latent} for SD1.5 and SDXL~\cite{podell2023sdxl} models and ADD-XL~\cite{sauer2023adversarial}. 
Note that ADD-XL is a GAN-based distillation method that generates exceptionally realistic samples but has evidence to provide poor image diversity for the given text prompt~\cite{sauer2023adversarial}.

We estimate the diversity of generated images by conducting a human study.
In more detail, given a text prompt and a pair of samples generated from different initial noise samples, assessors are instructed to evaluate the diversity of the following attributes: angle of view, background, main object and style.
For each model, the votes are collected for $600$ text prompts from COCO2014 and aggregated into the scores from 0 to 1, higher scores indicate more diverse images. 
The results are presented in Figure~\ref{fig:diversity_scores}.

\begin{figure*}
    \centering
    \includegraphics[width=0.6\linewidth]{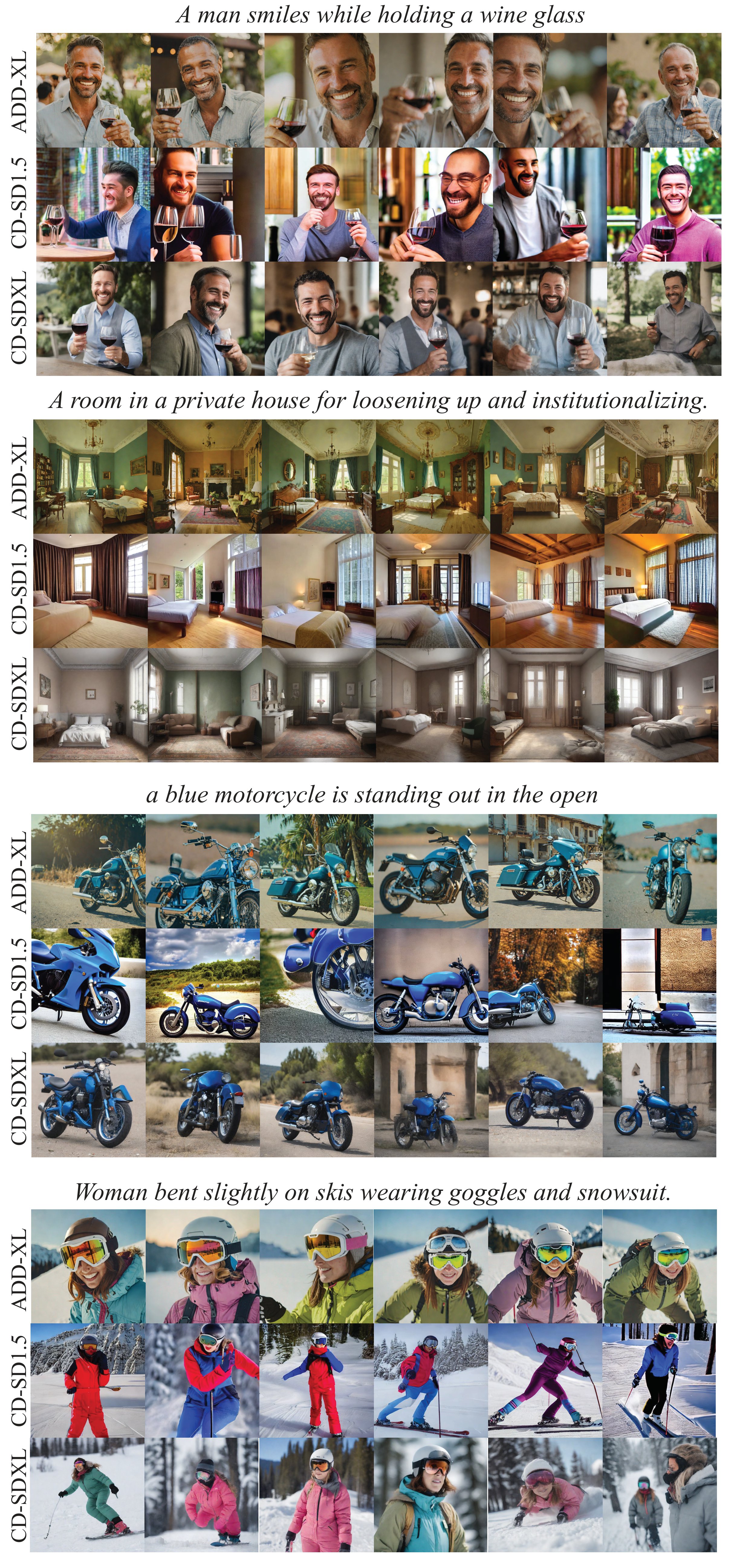}
    \caption{Visual examples generated with various distilled text-to-image models for different seed values. CD-based students generate more diverse images than ADD-XL.}
    \label{fig:diversity}
\end{figure*}

CD-SDXL demonstrates significantly better diversity than ADD-XL but still produces less various images compared to the SDXL teacher.
CD-SD1.5 performs similarly to the SD1.5 teacher.
Also, both adaptive strategies increase the diversity of the SDXL student models, especially the regeneration one.
%Notably, the adaptive approach with CD-based students demonstrates comparable results to the corresponding teacher models. 
In Figure \ref{fig:diversity}, we illustrate the diversity of images generated with different distilled text-to-image models (ADD-XL, CD-SDXL and CD-SD1.5). 
Each column corresponds to a different initial noise (seed). 
We notice that ADD-XL exhibits the lowest diversity compared to the CD-based counterparts.

Then, we address whether the distilled models tend to oversimplify the teacher distribution.
In this experiment, we evaluate SD1.5 using DDIM for $50$ steps and the corresponding CD-SD1.5 using $5$ sampling steps.
In~\fig{app_student_compl}, we compare the complexity of the student and teacher samples in terms of the ICNet score~\cite{feng2023ic9600}.
We observe that CD-SD1.5 imperceptibly simplifies the teacher distribution. %and, as a result, does not bring noticeable negative distribution changes.

To sum up, in our experiments, the CD-based models provide the decent distribution diversity that can be further improved with the proposed adaptive approach. 

\subsection{Controllable generation}
\label{app:applications}
For both tasks, we use the adaptive refinement strategy and set the rollback value $\sigma$ to $0.5$.
We perform $5$ steps for the student generation and $10$ steps for the refinement with the UniPC solver.
The cut-off thresholds correspond to $70$ and $50$ ImageReward percentiles for the mask-guided and edge-guided generation, respectively.
We select random $600$ image-text pairs from the COCO2014 validation set for the edge-guided generation.
For the mask-guided generation, we use $600$ semantic segmentation masks from the ADE20K dataset~\cite{ade20k} and use the category names as the text prompts.  
For evaluation, we conduct the human study similar to \ref{app:human_eval}.

\subsection{ImageReward inference costs}
\label{app:ir_inference_costs}
We compare the absolute inference times of a single Stable Diffusion UNet step with classifier-free guidance against the ImageReward forward pass.
We measure the model performance in half precision on a single NVIDIA A100 GPU.
The batch size is $200$ to ensure $100\%$ GPU utility for both models.
The performance is averaged over $100$ independent runs.
ImageReward demonstrates $0.26$s while the single step of Stable Diffusion takes $3$s.
In the result, we consider the adaptive step costs negligible since ImageReward is more than $10{\times}$ faster than a single generation step of Stable Diffision.

\subsection{Additional visualizations}
\begin{figure*}
    \centering
    \includegraphics[width=\linewidth]{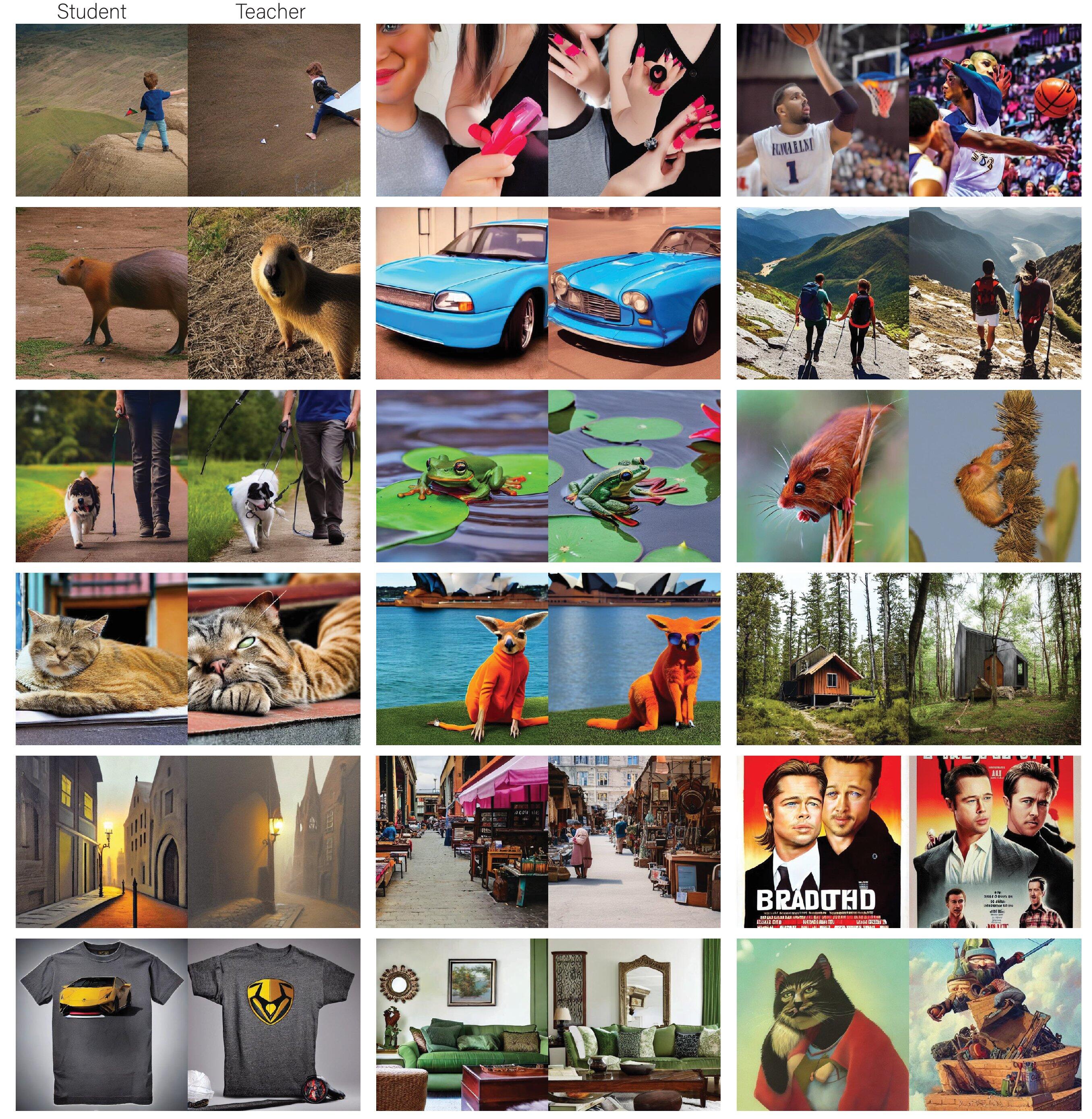}
    \caption{Additional examples where the student (CD-SD1.5) outperforms its teacher (SD1.5) according to the human evaluation.}
    \label{fig:app_student_better1}
\end{figure*}

\begin{figure*}
    \centering
    \includegraphics[width=\linewidth]{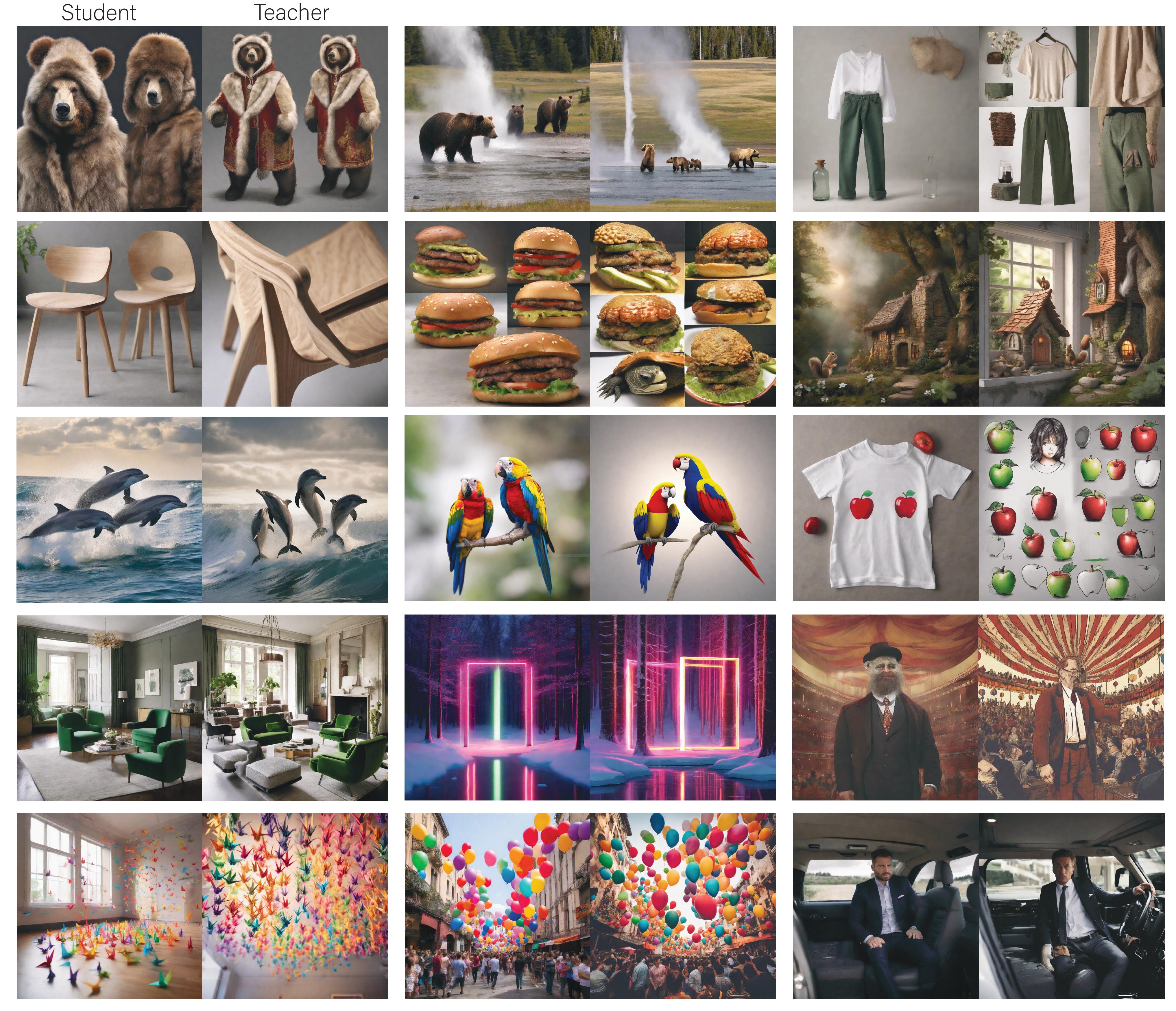}
    \caption{Additional examples where the student (CD-SDXL) outperforms its teacher (SDXL) according to the human evaluation.}
    \label{fig:app_student_better2}
\end{figure*}

\begin{figure*}
    \centering
    \includegraphics[width=\linewidth]{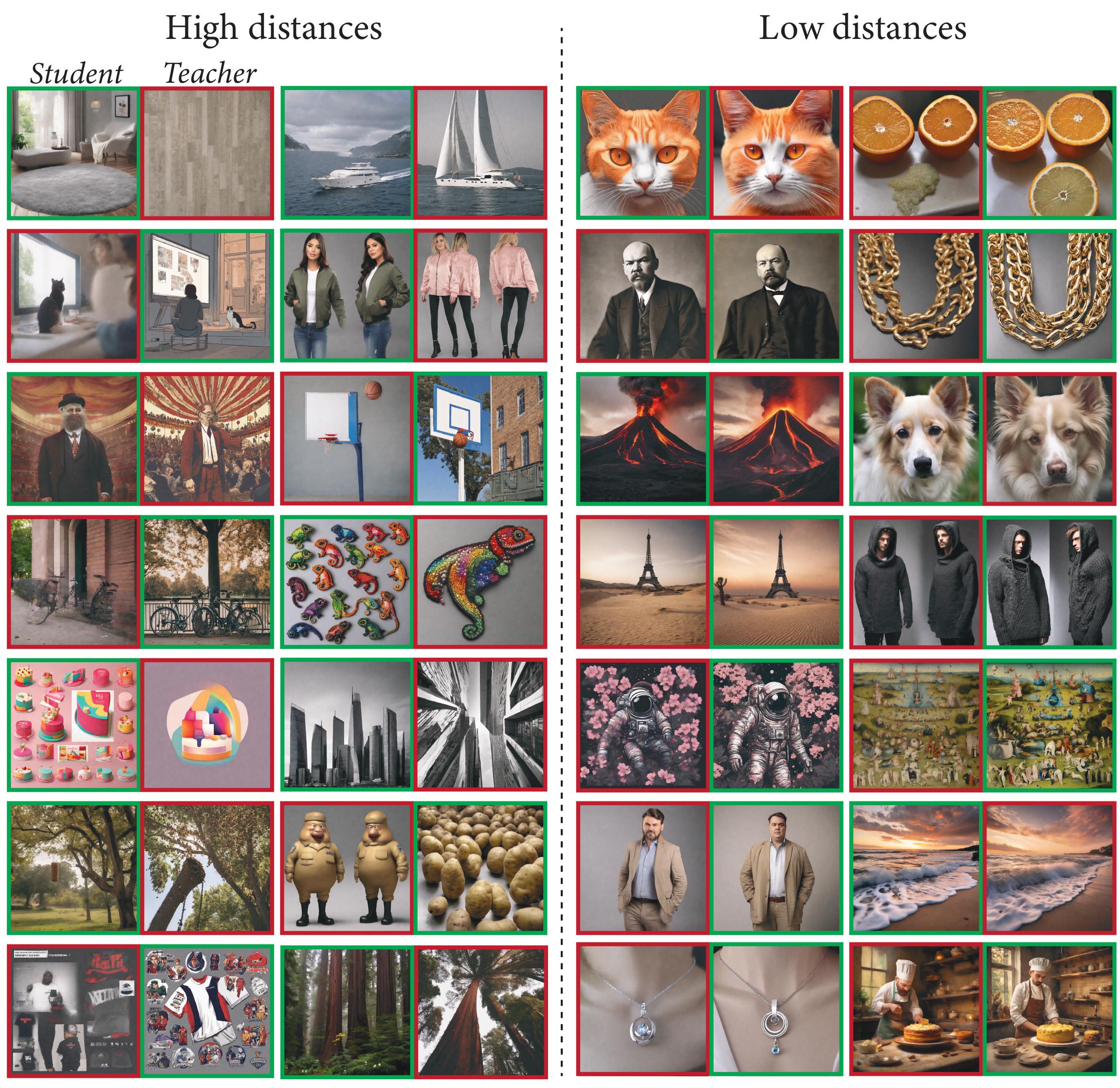}
    \caption{Visualization of the student (CD-SDXL) and teacher (SDXL) samples for low and high distance ranges. The green outline corresponds to wins, while the red one - losses.}
    % \caption{Qualitative verification that the student (CD-SDXL) wins are more likely where its samples differ from the teacher (SDXL) ones. The green outline means wins, while the red one - losses.}
    \label{fig:high_low_dist_better}
\end{figure*}

\begin{figure*}
    \centering
    \includegraphics[width=\linewidth]{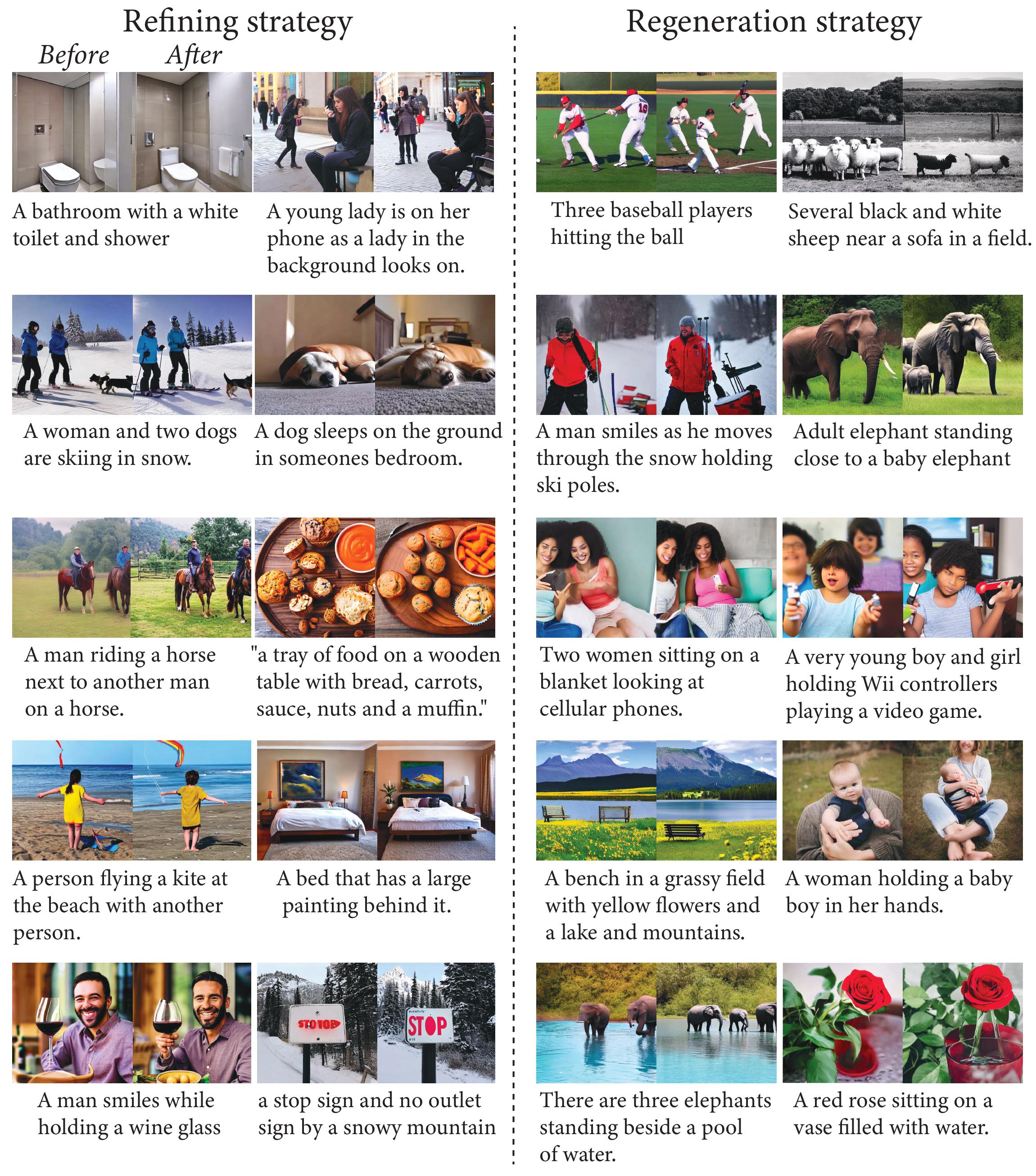}
    \caption{Visual examples of the refining and regeneration adaptive strategies.}
    \label{fig:refining_regen_strat}
\end{figure*}

In Figure~\ref{fig:app_student_better1}, \ref{fig:app_student_better2} we present qualitative verification of the first observation (the student sometimes outperforms its teacher according to the human evaluation). 
Figure~\ref{fig:high_low_dist_better} supports the second observation (the student wins are more likely where its samples differ from the teacher ones). 
In Figure~\ref{fig:refining_regen_strat} we demonstrate two adaptive strategies (refining and regeneration), the examples confirm that the refinement strategy improves the image fidelity and does not significantly alter the textual alignment, while the regeneration strategy may improve textual alignment.

\begin{figure*}
    \centering
    \includegraphics[width=\linewidth]{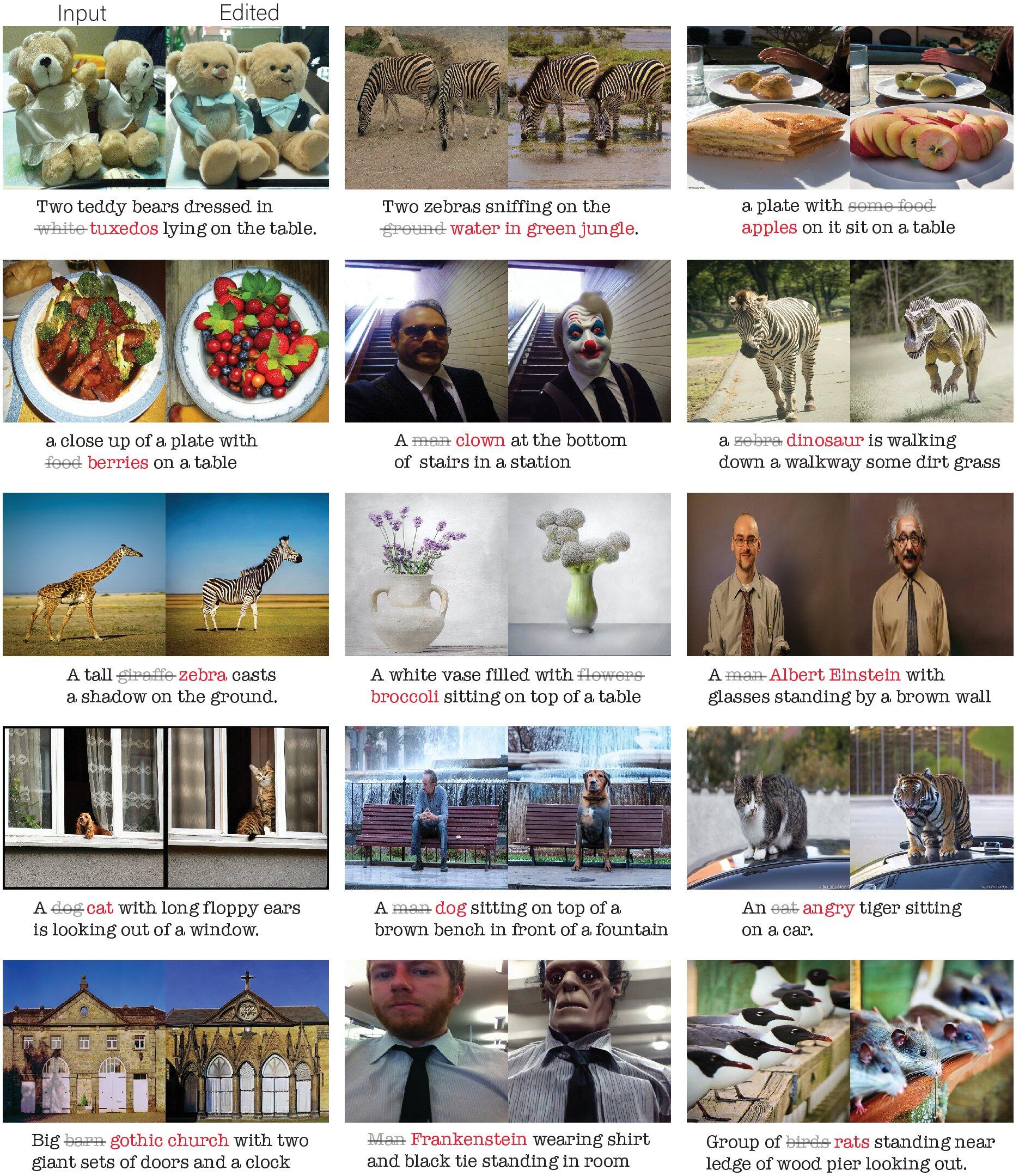}
    \caption{Additional image editing results produced with our approach.}
    \label{fig:app_editing_more}
\end{figure*}

\begin{figure*}
    \centering
    \includegraphics[width=\linewidth]{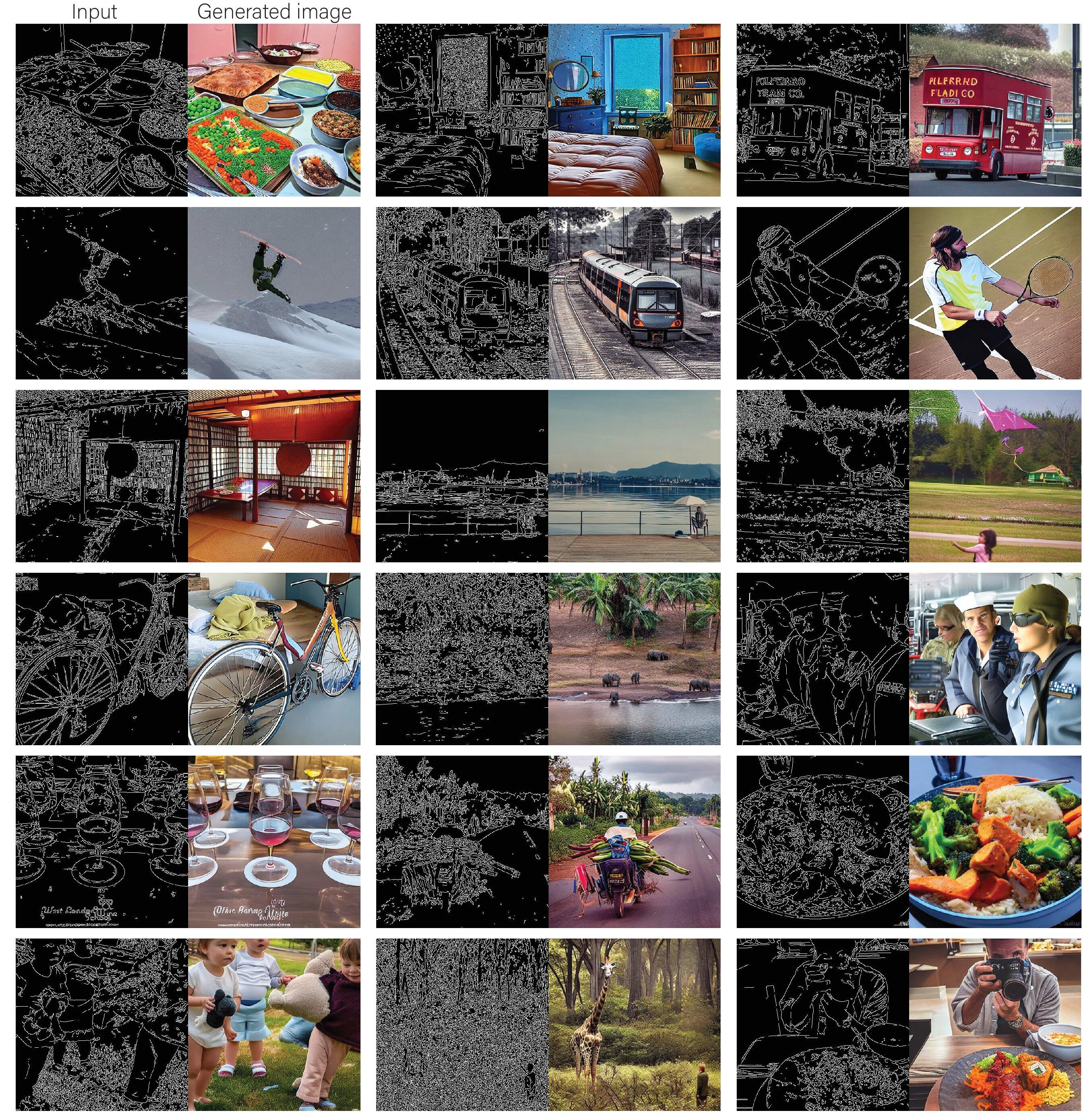}
    \caption{Additional results on Canny edge guided image generation with our approach.}
    \label{fig:app_canny_more}
\end{figure*}

\begin{figure*}
    \centering
    \includegraphics[width=\linewidth]{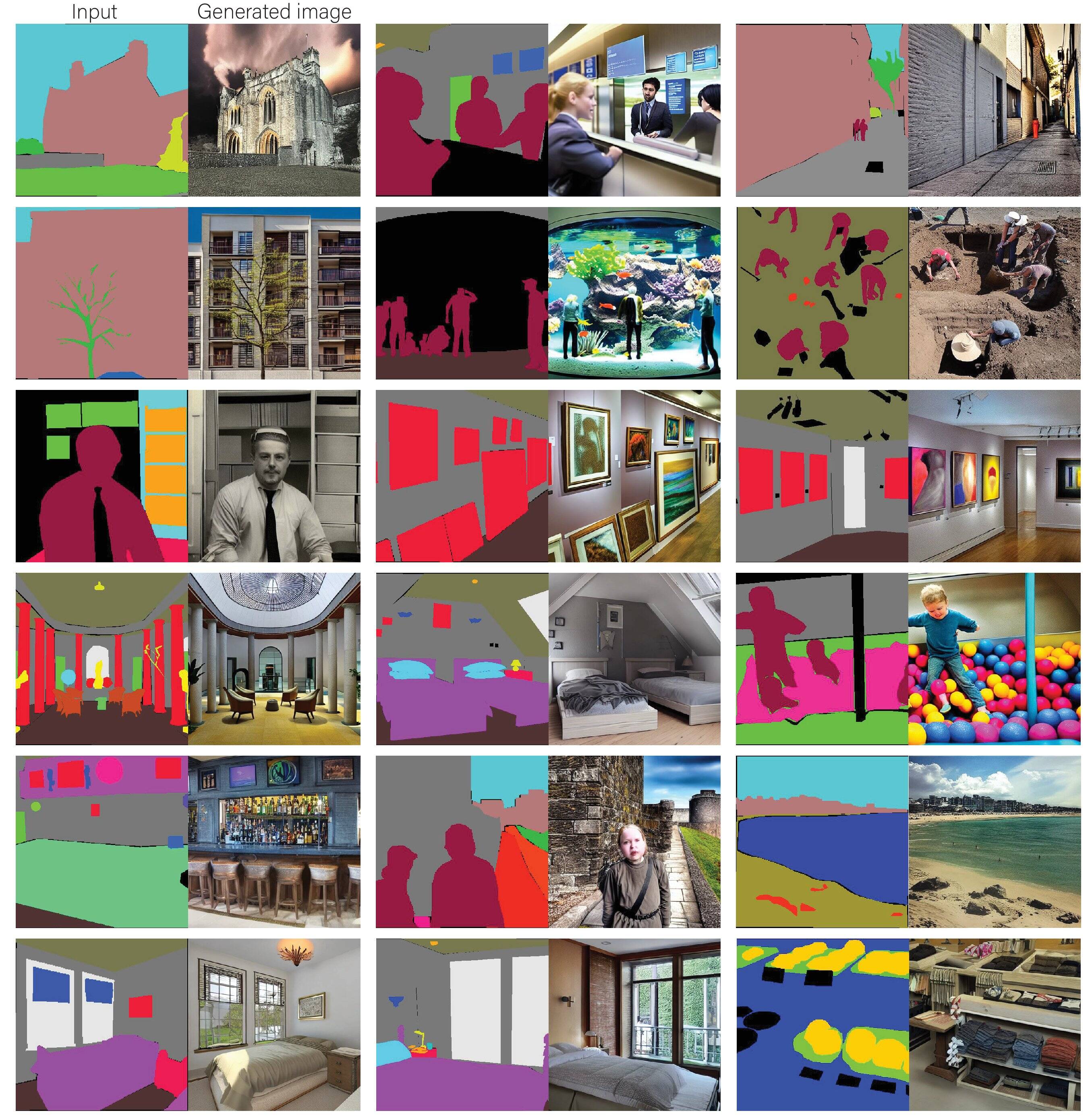}
    \caption{Additional results on segmentation mask guided image generation  with our approach.}
    \label{fig:app_segment_more}
\end{figure*}

Figures \ref{fig:app_editing_more}, \ref{fig:app_canny_more}, \ref{fig:app_segment_more} provide more qualitative comparisons of our approach for different tasks.

%\begin{figure*}
%    \centering
%    \includegraphics[width=\linewidth]{images/controllabel.pdf}
%    \caption{Student outperforms its teacher according to the human evaluation.}
%    \label{fig:app_control}
%\end{figure*}

% \section{Visualizations}
% In Figure \ref{fig:app_student_better1} we present  examples where the student outperforms its teacher according to the human evaluation. Moreover, 
% Figure \ref{fig:app_editing_more}, Figure \ref{fig:app_canny_more} and Figure \ref{fig:app_segment_more} demonstrates the qualitative results of our approach for different tasks.

\label{app:visualizations}